\documentclass[sigconf]{acmart}
\AtBeginDocument{%
  }

\copyrightyear{2024}
\acmYear{2024}
\setcopyright{rightsretained}
\acmConference[SIGGRAPH Conference Papers '24]{Special Interest Group on Computer Graphics and Interactive Techniques Conference Conference Papers '24}{July 27-August 1, 2024}{Denver, CO, USA}
\acmBooktitle{Special Interest Group on Computer Graphics and Interactive Techniques Conference Conference Papers '24 (SIGGRAPH Conference Papers '24), July 27-August 1, 2024, Denver, CO, USA}
\acmDOI{10.1145/3641519.3657522}
\acmISBN{979-8-4007-0525-0/24/07}

\usepackage{colortbl}
\usepackage{subcaption}
\usepackage{multirow}
\usepackage{tabularx}
\usepackage{amsmath,amsthm,enumitem}
\setlist{leftmargin=5.5mm}

\usepackage{soul}

\DeclareMathOperator*{\argmin}{arg\,min}

\begin{document}

\title{Towards Unstructured Unlabeled Optical Mocap: {\em A Video Helps!}}

 \acmSubmissionID{1253}

\author{Nicholas Milef}
\email{nicholas.milef@tamu.edu}
\affiliation{
    \institution{Texas A\&M University}
    \city{College Station}
    \state{Texas}
    \country{USA}
}

\author{John Keyser}
\email{keyser@cse.tamu.edu}
\affiliation{
    \institution{Texas A\&M University}
    \city{College Station}
    \state{Texas}
    \country{USA}
}

\author{Shu Kong}
\email{skong@um.edu.mo, shu@tamu.edu}
\authornote{Corresponding author.}
\affiliation{
    \institution{University of Macau}
    \city{Macau}
    \state{}
    \country{China}
}
\affiliation{
    \institution{Texas A\&M University}
    \city{College Station}
    \state{Texas}
    \country{USA}
}

\begin{teaserfigure}
    \centering
    \includegraphics[width=\linewidth]{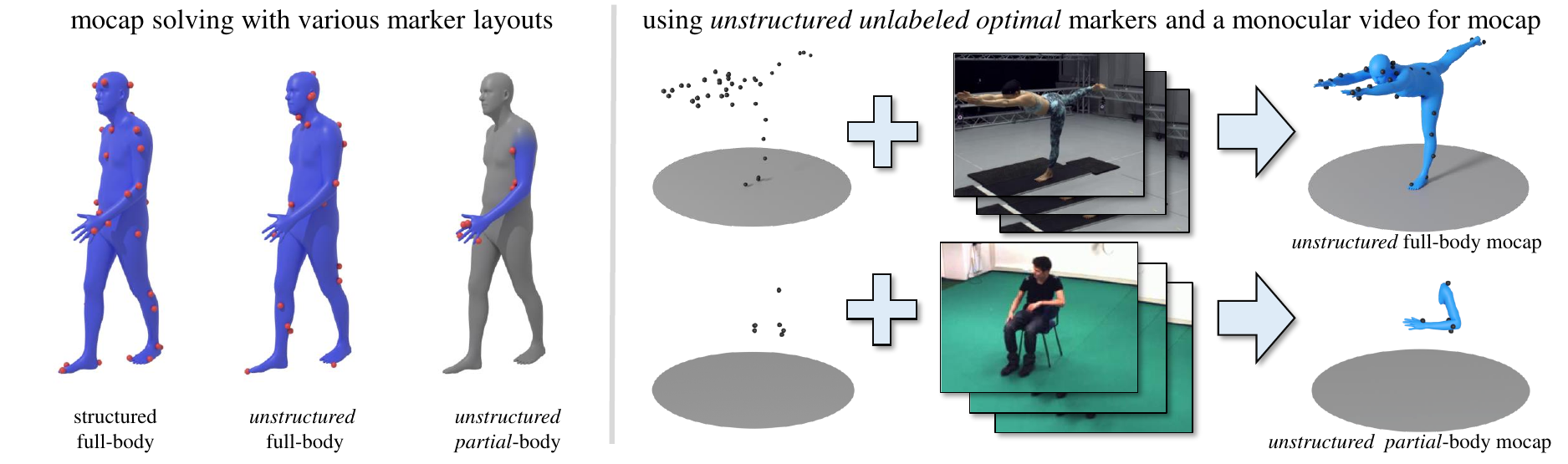}
    \caption{We solve the problem of unstructured unlabeled optical (UUO) motion caption (mocap), in which retroreflective optical markers are placed in an unstructured way on the body. 
    Importantly, markers are not manually labeled. UUO mocap reduces human effort to set up mocap environments but is more challenging than typical mocap settings that either manually label markers or place markers w.r.t some predefined structured layouts.
    To solve UUO mocap, we leverage a monocular video captured alongside markers and use it to extract an initial body model as a prior for subsequent optimization for body pose, shape, global translation, and rotation. 
    }
    \Description{We solve the problem of unstructured unlabeled optical (UUO) motion caption (mocap), in which retroreflective optical markers are placed in an unstructured way on the body. 
    Importantly, markers are not manually labeled. UUO mocap reduces human effort to set up mocap environments but is more challenging than typical mocap settings that either manually label markers or place markers w.r.t some predefined structured layouts.
    To solve UUO mocap, we leverage a monocular video captured alongside markers and use it to extract an initial body model as a prior for subsequent optimization for body pose, shape, global translation, and rotation. }
    \label{fig:teaser}
\end{teaserfigure}

\begin{abstract}
Optical motion capture (mocap) requires accurately reconstructing the human body from retroreflective markers, including pose and shape. In a typical mocap setting, marker labeling is an important but tedious and error-prone step. Previous work has shown that marker labeling can be automated by using a \textit{structured} template defining specific marker placements, but this places additional recording constraints. We propose to relax these constraints and solve for \textit{Unstructured Unlabeled Optical} (UUO) mocap. Compared to the typical mocap setting that either labels markers or places them w.r.t a structured layout, markers in UUO mocap can be placed anywhere on the body and even on one specific limb (e.g., right leg for biomechanics research), hence it is of more practical significance. It is also more challenging. To solve UUO mocap, we exploit a monocular video captured by a single RGB camera, which does not require camera calibration. On this video, we run an off-the-shelf method to reconstruct and track a human individual, giving strong visual priors of human body pose and shape. With both the video and UUO markers, we propose an optimization pipeline towards marker identification, marker labeling, human pose estimation, and human body reconstruction. Our technical novelties include multiple hypothesis testing to optimize global orientation, and marker localization and marker-part matching to better optimize for body surface. We conduct extensive experiments to quantitatively compare our method against state-of-the-art approaches, including marker-only mocap and video-only human body/shape reconstruction. Experiments demonstrate that our method resoundingly outperforms existing methods on three established benchmark datasets for both full-body and partial-body reconstruction.
\end{abstract}



\maketitle

\section{Introduction}
\label{ssec:introduction}
Human reconstruction is a crucial component for creating realistic humans in movies and games~\cite{holden2018robust, west2019going, bregler2007motion}, biomechanics analysis~\cite{averta2021u, camargo2021comprehensive, van2022biomechanics, MOESLUND200690, roetenberg2009xsens}, and computer vision applications~\cite{kocabas2020vibe, rempe2021humor, wang2023learning}.
This is a challenging problem as individuals have different body shapes and can express various poses. Optical motion capture (mocap) systems have been the de facto system to capture pose and body shape due to their high accuracy in determining 3D marker positions~\cite{merriaux2017study}. 
These systems use multi-view infrared cameras to recover the positions of retroreflective markers placed on the body.
Subsequently, one can fit a 3D body model (e.g., SMPL~\cite{SMPL:2015}) to the marker positions if the placement and corresponding body parts of markers are known~\cite{loper2014mosh, SMPL-X:2019}.

In optical mocap, accurate body reconstruction typically requires manually labeling markers and consistent placement~\cite{loper2014mosh, mahmood2019amass}.
When provided labeled markers, approaches such as HuMoR~\cite{rempe2021humor} and VPoser~\cite{SMPL-X:2019} can fit a 3D human body to marker locations by solving for pose and body shape through optimization. 
However, manually labeling markers is prone to errors and is time consuming; without labels for markers,
approaches like HuMoR and VPoser can fail to  reconstruct pose as optimization easily gets stuck on bad local minima.
Therefore, some methods propose to automate marker labeling~\cite{ghorbani2021soma, ghorbani2019auto}.
These methods are trained on some predetermined marker layouts and struggle to label markers placed w.r.t unseen layouts.
Such layouts could come from a more user-friendly setup that allows markers to be placed anywhere on the body, i.e. \textit{unstructured} mocap.
Finally, current approaches are not able to handle partial marker layouts such as markers placed on only the left leg or right shoulder.
\textit{Partial-body} reconstruction is critical for biomechanics research~\cite{averta2021u, camargo2021comprehensive, van2022biomechanics, MOESLUND200690, roetenberg2009xsens} which often seeks to minimize the number of unnecessary markers. 
However, it is difficult to precisely understand how markers cover the body part without marker labels or a full body reference.

We summarize the present dilemma of mocap: 
\emph{accurate labeling is crucial to accurate pose and body reconstruction, yet accurate labeling relies on accurate pose and body shape, which is challenging, if not impossible, with unstructured markers.}

This dilemma motivates our work of solving \textit{Unstructured Unlabeled Optical} (UUO) mocap, aiming for simultaneous human body reconstruction and pose estimation.
Inspired by recent advances in human reconstruction from monocular videos~\cite{goel2023humans},
we leverage a video captured by a commodity camera (such as a cellphone) along with UUO markers for mocap.
It is worth noting that the UUO mocap setup only requires the video to be temporally synchronized w.r.t optical markers.
It does not require (1) marker identification from video frames, and (2) camera calibration between the camera and multi-view infrared cameras in the mocap studio.
In other words, we exploit the monocular video to obtain a human body prior to aid mocap.
Hence, methods developed in this setup can generalize across a wide range of optical mocap systems.

We leverage the following insights to assist in combining UUO markers and the corresponding monocular video for solving mocap.
First,  modern pose estimation techniques from monocular video, though they struggle to predict global translation and absolute size, tend to produce relatively accurate poses and correctly estimate proportional body shape.
We therefore use such estimations to serve as pose priors for 3D model fitting. 
Second, part-based segmentation of markers, which assigns a body part label to each marker,
is relatively easy to solve.
We leverage a statistical human model together with the insight that body parts are relatively rigid to help find optimal marker fits. By finding motion correspondence between markers and video-estimated human motion, we can jointly label the markers and solve for human pose and shape.
Importantly, our method does not expect a structured marker layout; the markers can be placed anywhere or just on part of the body. Partial-body mocap is especially useful for biomechanics mocap~\cite{van2022biomechanics, averta2021u, camargo2021comprehensive} and animal mocap (where marker templates may not be available)~\cite{abson2015motion, zhang2018mode}.

We make three major contributions (cf. $\S$\ref{ssec:methodology}). 
\begin{itemize}[noitemsep,topsep=0pt]
    \item 
    {\bf Problem statement.}
    We introduce the problem of Unstructured Unlabeled Optical (UUO) mocap, aiming for simultaneous body reconstruction and pose estimation using UUO markers. It relaxes the constraints of marker placement and requires no manual work of labeling markers.

    \item {\bf New strategy.} 
    To solve UUO mocap, we leverage a monocular video captured alongside markers to extract a body prior using an off-the-shelf method of monocular human body reconstruction. We use this body prior for optimizing body/part pose, size, and marker locations.

    \item 
    {\bf Novel techniques.}
    We propose a UUO mocap pipeline consisting of multiple novel techniques such as
    (1) a multi-stage fitting process for temporally-stable motion reconstruction,
    (2)
    identifying and localizing markers by finding their correspondence to individual body parts, 
    and
    (3) multiple hypothesis testing for rotational alignment of body mesh.     
\end{itemize}

\noindent Code and data are at \href{https://github.com/NicholasMilef/UUO-Mocap}{https://github.com/NicholasMilef/UUO-Mocap}.

\section{Related Work}
\label{ssec:related_work}
\subsection{Statistical Human Models}
Mocap markers are placed near the surface of the skin, so one can use the marker 3D locations to model the body mesh. 
Various statistical methods propose to model the human body~\cite{SMPL:2015, SMPL-X:2019, xu2020ghum} by  introducing different vertex offsets from a based mesh template (e.g., blend shapes) that can be controlled through a learned parameter space.
SMPL~\cite{SMPL:2015} is a well-established statistical human model that has gained popularity in various applications,
The SMPL model is parameterized by body shape $\beta\in\mathbb{R}^{10}$, 
pose $\Theta\in\mathbb{R}^{23}$, 
global translation $\Gamma\in\mathbb{R}^{3}$, and global orientation $\Phi\in\mathbb{R}^3$.
The SMPL model is differentiable w.r.t vertex positions $V$ and joint positions $J$ defined as function $\mathcal{M}$:
$\left[J, V\right] = \mathcal{M}(\Phi, \Theta, \beta) + \Gamma$.
To solve for human pose and shape reconstruction from markers, various methods  fit an SMPL model to the labeled markers.
For example, 
VPoser~\cite{SMPL-X:2019}, a conditional variational autoencoder, learns a pose prior from the AMASS~\cite{mahmood2019amass} dataset and fits SMPL to labeled keypoints. HuMoR~\cite{rempe2021humor} extends this idea by using a motion prior to assist in keypoint fitting for both image and motion capture applications.
However, both of these approaches struggle to find strong initialization priors using just \emph{unlabeled} marker point clouds.
Our work also uses SMPL~\cite{SMPL:2015} to represent a 3D body but optimizes it for solving mocap and human body reconstruction in the \emph{UUO mocap setting}.

\begin{figure*}[t]
    \centering
    \includegraphics[width=\linewidth]{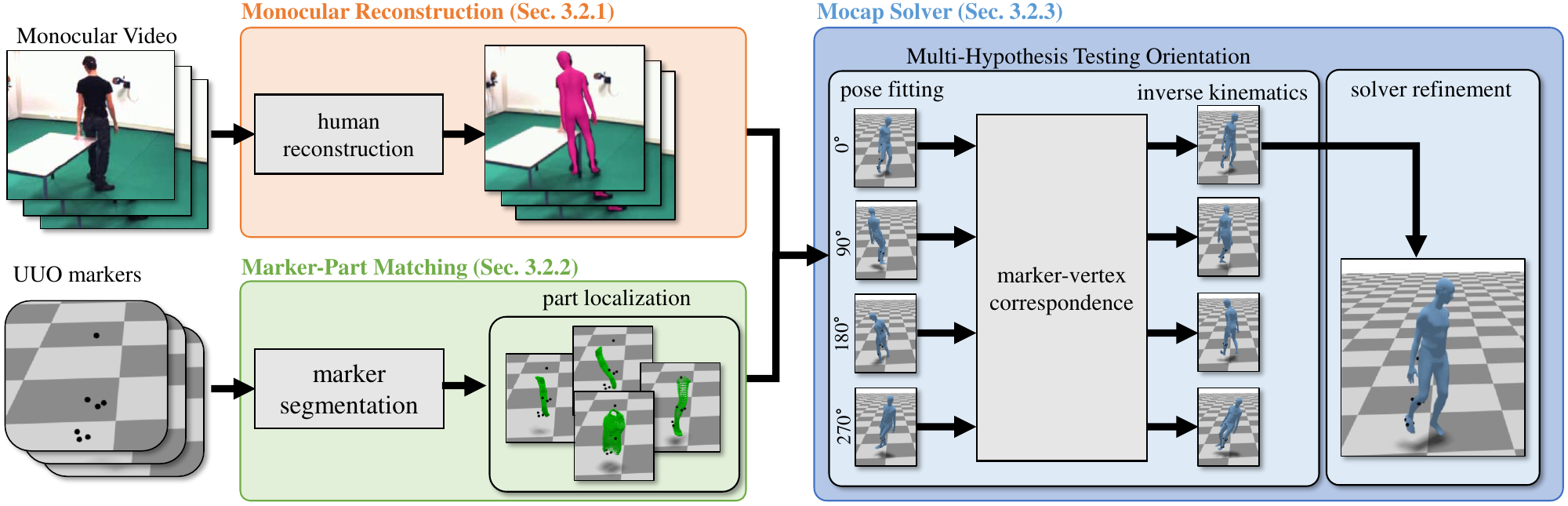}
    \caption{The proposed pipeline of our UUO mocap solver consists of three modules (cf. details in Section~\ref{ssec:methodology}).
    Our method takes as input monocular video and UUO markers to  jointly predict marker labels, pose, and body shape. First, we use an off-the-shelf method (HMR2.0~\cite{goel2023humans}) to generate a human prior from the video. Then, we segment the 3D mocap markers to estimate the number of bones that need to be reconstructed. Then, we search for the best-fitting body part. Finally, we solve for the pose and body shape through a novel optimization process.}
    \label{fig:system}
    \Description{The proposed pipeline of our UUO mocap solver consists of three modules (cf. details in Section~\ref{ssec:methodology}).
    Our method takes as input monocular video and UUO markers to  jointly predict marker labels, pose, and body shape. First, we use an off-the-shelf method (HMR2.0~\cite{goel2023humans}) to generate a human prior from the video. Then, we segment the 3D mocap markers to estimate the number of bones that need to be reconstructed. Then, we search for the best-fitting body part. Finally, we solve for the pose and body shape through a novel optimization process.}
\end{figure*}

\subsection{Motion Capture Solving}
Motion capture solving typically uses {\em labeled} markers to determine body pose and/or shape
via optimization~\cite{loper2014mosh, mahmood2019amass}, 
which fits a body model to markers by minimizing distances between labeled markers and vertices of the body model. 
Recent approaches perform mocap solving and marker denoising via deep learning~\cite{han2018online, chen2021mocap, pan2023locality}.
These approaches assume the marker inputs to be labeled already.
Some prior works attempt to mitigate the need of marker labeling.
Han et al.~\shortcite{han2018online} use a deep neural network and bipartite matching to label mocap markers placed on a hand and assume a structured marker layout. 
Holden~\shortcite{holden2018robust} uses a residual network to jointly denoise marker positions and solve for pose for each frame. MocapSolver~\cite{chen2021mocap} estimates marker offsets from each skeletal joint, bone lengths, and the pose using a window of frames consisting of mocap marker positions.
Other follow-up works adopt deep learning and optimization to improve mocap using unlabeled markers~\cite{tang2023divide, pan2023locality}.
However, these methods all require a known marker layout.
Our work distinguishes from existing ones in that we, for the first time, solve mocap using \emph{unstructure unlabeled optical (UUO) markers}.

\subsection{Automatic Marker Labeling}
Traditional mocap workflows require technicians to manually label markers, which is time-consuming and error-prone.
Hence, some works study marker auto-labeling.
Among plenty of marker auto-labeling and mocap solvers,
Meyer et al.~\shortcite{meyer2014online} propose an online labeling solution, but they require the actor to perform a T-pose for initialization, making it unsuitable for archival data. 
Schubert et al.~\shortcite{schubert2015automatic} propose an automatic mocap solver and marker location finder with a reasonably dense marker layout. However, their method needs a database of human pose/shape templates, limiting the method to poses and shapes present in the database.
Alexanderson et al.~\shortcite{alexanderson2017real} propose an algorithm that solves unlabeled markers on the hands and head but does not temporally lock the marker labels and requires knowledge of the orientation of these body parts before running.
Recent works propose to train neural networks on a database of defined human pose/shape layouts and use the trained networks to automatically label markers and solve mocap~\cite{ghorbani2019auto, clouthier2021development, ghorbani2021soma}.
For example, SOMA~\cite{ghorbani2021soma} labels unlabeled mocap markers through a per-frame self-attention network. After labeling, the markers can then be used to optimize for body shape and pose using MoSh++~\cite{mahmood2019amass}. While SOMA trains a ``SuperSet'' that can work for a variety of known marker layouts, it does not generalize to unseen or partial-body layouts.
In sum, existing approaches require a database of defined layouts to train networks for marker labeling and mocap, limiting their generalization to new marker layouts. 
Our problem of UUO mocap does not provide marker labels and requests the study of mocap with unstructured markers, hence solutions to this problem is of practical significance in mocap systems.

\subsection{Monocular Video Mocap} 
Markerless mocap has become a popular alternative to traditional optical marker-based mocap systems.
Mocap from monocular RGB video is the most accessible form of markerless mocap due to the ubiquity of RGB cameras. 
For monocular video mocap, recent methods adopt model-based optimization~\cite{Bogo:ECCV:2016, SMPL-X:2019, rempe2021humor} and deep learning~\cite{kanazawa2018end, zhang2021pymaf, zhang2023pymaf, goel2023humans}, resulting in reasonably accurate poses and proportional body shape (e.g., absolute measures such as height may be inaccurate). However, recovering global position and rotation in the world is still difficult during monocular reconstruction~\cite{yuan2022glamr, ye2023decoupling}.
Some approaches have sought to augment different forms of tracking data. One popular form of mocap, due to cheap cost and less setup, is Inertial Measurement Unit (IMU) based motion capture. Combining video and IMU data has been shown to be more effective than just using video or IMU data alone~\cite{tan2022imu, pearl2023fusion}.  The combination even approaches optical mocap accuracy \cite{shin2023markerless}. 
However, IMU sensors have a tendency to drift that video cannot fully fix, causing lower accuracy~\cite{van2018accuracy}. Another approach has been to use depth maps with optical mocap marker positions. Some approaches~\cite{chatzitofis2021democap, chatzitofis2022low} use a few low-cost multi-view depth cameras to jointly label markers and solve for pose but theydo not match the performance of professional optical mocap systems.
In our work, we show that monocular video can {\em assist} in optical marker-based mocap. Experiments demonstrate that our mocap solver, which exploits both UUO markers and the corresponding monocular video, outperforms methods based on either, approaching the performance of a labelled-marker based solver.

\section{Problem Definition and Methodology}
\label{ssec:methodology}
We first present the formal problem definition of Unstructured Unlabeled Optical (UUO) mocap, then explain our method for solving UUO mocap, and finally present important implementation details.

\subsection{Problem Definition}
The problem of \emph{unlabeled and unstructured optical (UUO) mocap} aims to reconstruct a full/partial body pose and shape from a sequence of \emph{unlabeled} markers, which are placed without a predefined structure on an individual's body or body part.
The problem relaxes some unfriendly constraints in existing mocap: (1) it does not require manual labeling for the markers (so alleviating manual intervention), and (2) it does not require placing markers on a predefined layout (so reducing human effort in mocap setup and allowing reconstructing a partial body). While solutions to UUO mocap are of practical significance in mocap,
solving mocap under the UUO setting is more challenging than existing settings due to the lack of marker labels and any predefined structured marker placement.
Next, we elaborate our method for UUO mocap.

\subsection{The Proposed UUO Mocap Method}

The core insight of our method is using a monocular video to reconstruct human body as a prior for subsequent mocap solving.
Our method takes as input the sequence of UUO markers and the corresponding video, and maps them into SMPL~\cite{SMPL:2015} parameters.
We denote $M$ markers from $T$ frames as a set of 3D points by $m \in \mathbb{R}^{T\times |M|\times 3}$.
Fig.~\ref{fig:system} sketches the pipeline of our method, consisting of three modules.
First, {\em monocular reconstruction} produces an initial SMPL model from the video that extracts pose, shape, and relative rotation across time (\S \ref{ssec:video-processing}).
Second, 
{\em marker-part matching} finds the best correspondences between UUO markers and the initial SMPL model (\S \ref{ssec:marker-SegLoc}). 
Third, {\em mocap solver} takes the output of the marker-part correspondence and optimizes for the final result (\S \ref{ssec:mocap-solving}).
We elaborate them in the following three subsections.

\subsection{Monocular Reconstruction from Video}
\label{ssec:video-processing}

Monocular video contains important visual cues that are not present in raw marker data and can serve as a strong prior for marker fitting.
Our strategy of exploiting this video is to estimate an initial body from it.
Human reconstruction from  monocular video is a well-studied area and various methods have been proposed in the literature. 
In this work, we use the state-of-the-art method HMR2.0~\cite{goel2023humans} as an off-the-shelf method, which returns parameters of the well-established SMPL model~\cite{SMPL:2015}, e.g., pose $\Theta$ and shape $\beta$.
It is worth noting that, while SMPL is widely used in mocap from (labeled) markers, we make the first attempt of using it to represent a body prior extracted from a monocular video for UUO mocap.
As a monocular video has ambiguities in body size, orientation, and world translation, the SMPL model estimated from it unlikely aligns with real marker positions. 
Therefore, we adopt the next modules to incorporate it for mocap solving.

\subsection{Marker-Part Matching}
\label{ssec:marker-SegLoc}
Mocap essentially requires finding the correspondence of markers and body (or vertices of body surface).
With the initial SMPL model estimated from the monocular video in the previous module, intuitively one can  optimize this SMPL model by fitting it to the UUO markers.
However, directly solving this optimization problem can easily get stuck to bad local minima. 
Therefore, we aim to provide a better initialization for mocap  solving with a marker-part matching module,
which finds correspondences of markers and body parts.
This module is not only important for full-body mocap but also particularly crucial for partial-body mocap, because without marker-body correspondence, it is difficult, if not impossible, to find a good matched body part for markers without signals from the full body (cf. Table~\ref{tab:comparisons_markers_partial}).

In Marker-Part Matching, we first use the median marker position to align the SMPL mesh.
This provides a good initialization of global translation (for both full-body and partial-body mocap).
Next, we adopt a two-step process to find marker-part correspondence.

\subsubsection{Step 1: marker segmentation}
Recall that in SMPL, each vertex $v$ on the body surface has an associated linear blend skinning (LBS) weight that is the summation of the bones. 
The maximal LBS weight for the vertex can be used to indicate which bones the vertex belongs to; vertices belonging to the same bones are approximately related by a rigid transformation.
Therefore, finding marker-bone correspondence largely simplifies mocap solving for body part reconstruction.
But searching for the best correspondence requires testing all combinations of markers and SMPL bones and hence computationally expensive.
Therefore,
to reduce the search space,
we propose to group markers and presume each group is corresponding to a specific bone.
For example, for partial-body mocap of a leg (i.e., thigh, calf, foot), we would only be interested in searching for kinematic chains consisting of 3 bones.
Inspired by marker clustering~\cite{de2006automatic}, we use the standard deviation of the Euclidean distance between every pair of markers across all frames in the sequence. 
Note that one can easily obtain marker-marker correspondence across time by tracking them.
Then we construct an affinity matrix that consists of these standard deviations.
Lastly we use agglomerative clustering with average linkage~\cite{scikit-learn} and distance threshold of 5mm,
resulting into $K$ segmented groups of markers for each timestamp.

\subsubsection{Step 2: multiple hypothesis testing for part localization}
We adopt a search-based method to determine where the markers are located on the body.
Note that the $K$ groups of segmented markers can be interpreted as a kinematic chain $\mathcal{S}$ that contains a group of $K$ bones.
Therefore, we aim to find the best match between the $K$ groups of markers and a kinematic chain from a pool of $K$-bone candidate chains generated from the initial SMPL. 
For the pool of candidate chains, 
we extract all possible chains with length $K$ from the hierarchy of SMPL bones.
Then, for each of the candidate chain $\mathcal{S}$, whose vertices are denoted as $V_{\mathcal{S}}\subset V$,  
we fit the vertices of $V_{\mathcal{S}}$ to all the markers $M$.
Concretely, for a marker $m$ at time-$t$ (i.e., $m^{(t)} \in M^{(t)}$),
whose corresponding part of the candidate chain $\mathcal{S}$ at time-$t$ is denoted as $V^{(t)}$, we find the closest vertex $v^{(t)} \in V^{(t)}$.
Lastly, we select the candidate chain that produces the minimum fitting error, which is defined as $E_{\mathcal{S}} = \lambda_{\overrightarrow{\text{3D}}} E_{\overrightarrow{\text{3D}}} + \lambda_{\beta} E_{\beta}$ that consists of two terms explained below.
$E_{\overrightarrow{\text{3D}}}$ is a single-directional Chamfer distance loss between markers and vertices of $V_{\mathcal{S}}$:
\begin{equation} \label{eq:chamfer_unidirectional}
    E_{\overrightarrow{\text{3D}}} = \frac{1}{|T|\cdot|M|}\sum^{T}_{t=1}\sum_{m^{(t)}\in M^{(t)}}{\min_{v^{(t)}\in{V^{(t)}}}{\lVert v^{(t)} - m^{(t)} \rVert^2}}
\end{equation}
$E_{\beta}$ is a mean-squared error that regularizes the body shape against the initial SMPL body shape $\beta^{\text{img}}$:
\begin{equation}
    E_{\beta} = (\hat{\beta} - \beta^{\text{img}})^2 / |\beta^{\text{img}}|
\label{eq:shape-MSE}
\end{equation}

The number of chain candidates increases with the number of bones $K$. 
Searching over a large number of chains, e.g., for full-body reconstruction, can be computationally expensive.
Fortunately, for full-body reconstruction, the exact kinematic chain selection is less relevant as our system can generally recover from poor initialization in later stages.
Moreover, we reduce searching computation by removing redundant candidate chains which contain $\ge$90$\%$ of the same bones to other candidates.
Running this module produces correspondences of UUO markers to vertices of the initial SMPL model,
allowing for the subsequent mocap solving towards body pose and shape reconstruction.

\subsection{Mocap Solving}
\label{ssec:mocap-solving}

Given UUO markers positions and the SMPL model (containing parameters of body shape $\beta$, pose $\Theta$, global translation $\Gamma$, and global orientation $\Phi$ from the previous two modules),
we solve mocap by fitting the SMPL model to the mocap marker positions.
Intuitively, one can optimize them altogether, but the solution is easily stuck in bad local minima due to the difficulty of optimizing rotation $\Phi$.
Therefore, we break up the mocap solving process into a sequence of optimization stages and propose a multiple hypothesis testing for solving rotation.

\subsubsection{Stage 1: multiple hypothesis testing for root rotation}
We first predefine a grid of initialized rotational offsets. 
We optimize each of them alongside the rest of the SMPL parameters. Then, we select the SMPL parameters from the best-fitting rotational offset.
Our work assumes that the global rotations of the initial SMPL meshes and markers differ by a yaw rotation $b$ such that $\Phi_{\text{img}} = b\times\hat{\Phi}$. 
To facilitate rotation optimization, we independently optimize the rotation offset $b$, initialized by four uniformly-distributed values $y \in \mathcal{B} =\{0^{\circ}, 90^{\circ}, 180^{\circ}, 270^{\circ}\}$ using Stages 2 to 4 (described next). 
We then select the best optimized rotations such that:
\begin{equation}
    \beta, \Theta, \Gamma, \Phi = \argmin_{y\in \mathcal{B}}E_{\overrightarrow{\text{3D}}}(\beta^y,\Theta^y,\Gamma^y,\Phi^y)
\end{equation}

\subsubsection{Stage 2: pose fitting} 
With an initial rotational offset $y \in \mathcal{B}$,
we optimize $b$ along with pose $\Theta$, translation $\Gamma$, and shape $\beta$ by minimizing the following:
\begin{align}
    E_{\text{pose}} = \lambda_{\overrightarrow{\text{3D}}} E_{\overrightarrow{\text{3D}}} + \lambda_{\beta} E_{\beta} + \lambda_{\Theta} E_{\Theta}
\end{align}
where $E_{\overrightarrow{\text{3D}}}$ is defined in Eq.~\ref{eq:chamfer_unidirectional},
$E_{\overrightarrow{\beta}}$ is defined in Eq.~\ref{eq:shape-MSE}, and $E_{\Theta}$ regularizes pose which is defined below:
\begin{align}
     E_{\Theta} = \frac{1}{|T|\cdot|\Theta|}\sum_{t=1}^{T}(\Theta^{\text{img}}_t - \hat{\Theta}_t)^{2}
\end{align}

\begin{figure}
    \centering
    \includegraphics[width=\linewidth]{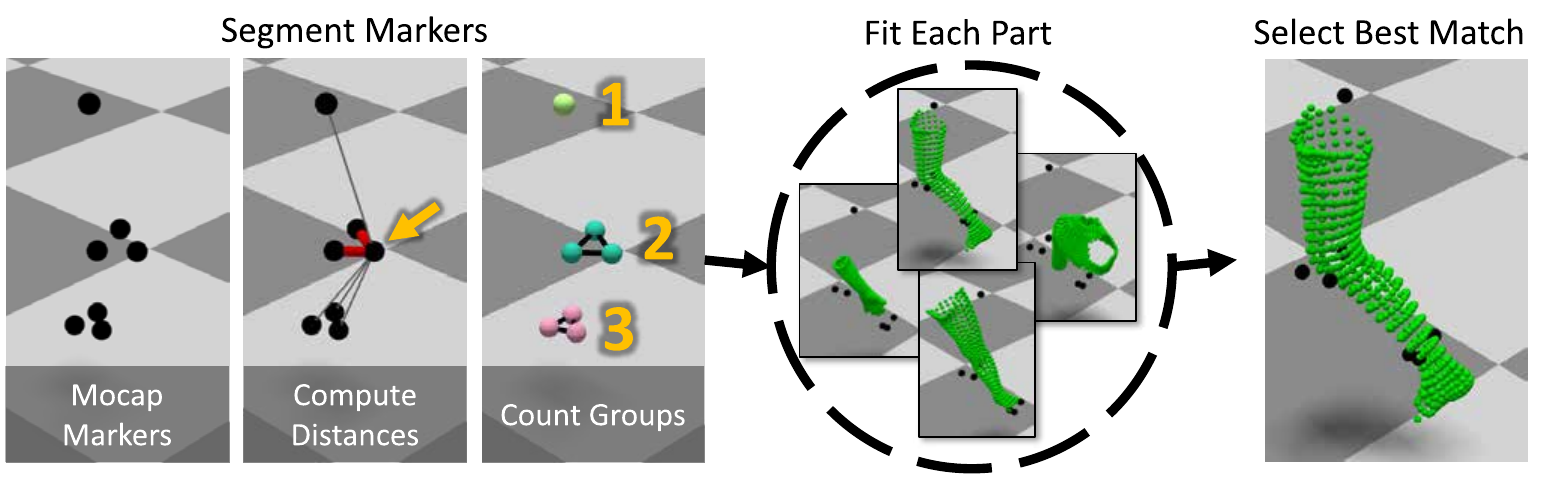}
    \\
    \caption{Our Marker-Part Matching first computes the standard deviation of distances between every other marker across all frames, then uses them to construct an affinity matrix to clustering markers into groups,
    and conducts hypothesis testing to select the best match that produces the minimum fitting error w.r.t the initial body model obtained from the monocular video.}
    \Description{Our Marker-Part Matching first computes the standard deviation of distances between every other marker across all frames, then uses them to construct an affinity matrix to clustering markers into groups,
    and conducts hypothesis testing to select the best match that produces the minimum fitting error w.r.t the initial body model obtained from the monocular video.}
    \label{fig:segmentation}
\end{figure}

\subsubsection{Stage 3: marker-vertex correspondence}
To better reconstruct the human body, 
for each marker $m$, we find the vertex $v^m$ of the optimized SMPL mesh with the closest average distance to $m$ over the entire sequence:
\begin{align}
    v^{m} = \argmin_{v \in V}{(\frac{1}{|T|}\sum^{T}_{t=1}{\Vert v^{(t)} - m^{(t)}\Vert_2})}
\end{align}
where $v^{(t)}$ and $m^{(t)}$ denote the vertex and marker at time-$t$, respectively.
With this found marker-vertex correspondence, we adopt inverse kinematics below that refines body reconstruction.

\subsubsection{Stage 4: inverse kinematics}
With the identified marker locations on the mesh surface from Stage 3, we solve an inverse kinematics (IK) problem via optimizing pose, shape and mesh:
\begin{equation}
    E_{\text{IK}} = \lambda_{M} E_{M} + \lambda_{\beta} E_{\beta} + \lambda_{\Theta} E_{\Theta}
\end{equation}
where minimizing $E_M$ will better align the SMPL mesh with markers. $E_M$ is a squared L2 norm loss between each marker $m^{(t)}$ and its corresponding vertex position $v_m$:
\begin{equation}
    E_M = \frac{1}{|T|\cdot|M|}\sum^{T}_{t=1}\sum_{m^{(t)}\in M}{(\lVert v^{(t)}_m - m^{(t)} \rVert_2 - \delta)^{2}}
\end{equation}
We set $\delta=9.5$ (in mm) which is a common mocap marker offset from the skin~\cite{ghorbani2021soma}.

\subsubsection{Stage 5: solver refinement}
Finally, we repeat the stages 3 and 4 one time to refine body reconstruction.
Instead of regularizing against the initial SMPL model $\Theta^\text{img}$, we regularize the pose against the output from Stage 4.
This helps to reduce distance inconsistencies between markers and the corresponding vertices.
We do not repeat more times as more iterations do not yield notable improvements despite more computation.

\subsection{Implementation}
For our mocap solver, we set the learning rate of the L-BFGS solver as 1.0 for part localization (Step 2) and inverse kinematics (Stage 4) and 0.1 for pose fitting (Stage 2).
We use terminal tolerances of 1e-7 on first order optimality and 1e-9 on function value/parameter changes. 
We process the entire sequence at once and optimize for a maximum of 10k iterations.
We tune these hyperparameters on a few random annotated examples (e.g., from UMPM).
After tuning, we use the same hyperparameters throughout our experiments for all the datasets (including both full-body and partial-body mocap):
$\lambda_{\overrightarrow{\text{3D}}}=10$ and $\lambda_{\beta}=0.1$ for Step 2;
$\lambda_{\Theta}=1$, $\lambda_{\overrightarrow{\text{3D}}}=10$, $\lambda_{\beta}=1$ at Stage 2;
$\lambda_{M}=1$, 
$\lambda_{\Theta}=0.1$, $\lambda_{\beta}=1$ at Stage 4.

\section{Experiments}
\label{ssec:experiments}

{
\setlength{\tabcolsep}{0.56em}
\begin{table*}[t]
    \centering
    \small
    \caption{\small
    Comparison of different methods for full-body reconstruction with UUO markers on three datasets.
    HMR2.0 is a method of reconstructing human body from a monocular video, we use rigid registration w.r.t UUO markers for its output (i.e., HMR2.0+RR) as a modified version that can serve UUO mocap. It underperforms the marker-only method SOMA.
    SOMA trains on marker layouts similar to CMU Kitchen so that it yields better numeric metrics on this dataset than the other two.
    Nevertheless, our method resoundingly outperforms all the compared approaches,
    approaching the performance of the reference which uses labeled markers.
    }
    \begin{tabular}{llcccccccccccc}
        \toprule
        \multirow{2}{*}{Method} & \multirow{2}{*}{Modality} & \multicolumn{4}{c}{UMPM} & \multicolumn{4}{c}{MOYO} & \multicolumn{4}{c}{CMU Kitchen} \\
        \cmidrule(lr){3-6}
        \cmidrule(lr){7-10}
        \cmidrule(lr){11-14}
        & & m2s$\downarrow$ & MPJPE$\downarrow$ & MPJVE$\downarrow$ & V2V$\downarrow$ & m2s$\downarrow$ & MPJPE$\downarrow$ & MPJVE$\downarrow$ & V2V$\downarrow$ & m2s$\downarrow$ & MPJPE$\downarrow$ & MPJVE$\downarrow$ & V2V$\downarrow$ \\
        \midrule
        VPoser & markers & 204.9 & 713.5 & 2962.1 & 738.7 & 39.3 & 612.4 & 1892.0 & 638.2 & 40.4 & 371.4 & 857.9 & 394.5 \\
        HuMoR & markers & 195.3 & 651.4 & 2464.8 & 689.5 & 42.3 & 607.9 & 1828.0 & 636.2 & 44.5 & 369.3 & 873.4 & 395.6 \\
        SOMA & markers & 26.8 & 101.1 & 151.5 & 95.5 & 54.3 & 268.5 & 102.9 & 276.2 & 17.0 & 88.0 & \textbf{23.7} & 90.9 \\
        \midrule
        HMR2.0+RR & markers+video & 150.1 & 334.1 & 515.1 & 360.7 & 146.6 & 430.5 & 305.8 & 448.3 & 131.9 & 396.3 & 267.8 & 425.4 \\
        VPoser+V & markers+video & 201.5 & 524.4 & 3134.6 & 572.5 & 19.2 & 132.2 & 1299.1 & 150.1 & 33.9 & 321.2 & 436.3 & 404.9 \\
        HuMoR+V & markers+video & 249.7 & 558.1 & 2137.6 & 598.2 & 32.5 & 205.9 & 1259.7 & 234.3 & 33.2 & 308.7 & 316.3 & 376.6 \\
        {\bf our method} & markers+video & \textbf{11.0} & \textbf{60.8} & \textbf{81.5} & \textbf{62.6} & \textbf{15.5} & \textbf{65.2} & \textbf{37.9} & \textbf{78.1} & \textbf{10.6} & \textbf{44.2} & 27.2 & \textbf{47.4} \\
        \midrule
        \cellcolor{gray!15}Reference & \cellcolor{gray!15}labeled markers & \cellcolor{gray!15}9.7 & \cellcolor{gray!15}0.00 & \cellcolor{gray!15}0.00 & \cellcolor{gray!15}0.00 & \cellcolor{gray!15}27.9 & \cellcolor{gray!15}0.0 & \cellcolor{gray!15}0.0 & \cellcolor{gray!15}0.0 & \cellcolor{gray!15}11.1 & \cellcolor{gray!15}0.0 & \cellcolor{gray!15}0.0 & \cellcolor{gray!15}0.0 \\
        \bottomrule
    \end{tabular}
    \label{tab:comparisons_markers_full}
\end{table*}
}

We conduct extensive experiments to validate our method by comparing against prior art. We also show rigorous ablation studies to demonstrate the importance of each step in our proposed optimization pipeline.
We start with setups of datasets and metrics, followed by  comprehensive results with in-depth analyses and discussions.

{
\setlength{\tabcolsep}{1em}
\begin{table*}
    \centering
    \small
    \caption{\small
    Comparison of different methods for partial-body reconstruction with UUO markers on the UMPM dataset~\cite{HICV11:UMPM}. 
    Different from the conclusions in full-body reconstruction (Table \ref{tab:comparisons_markers_full}), SOMA, as well as other marker-only methods, underperforms HRM2.0+RR, showing the limitation of marker-only mocap approaches for partial-body reconstruction.
    Our method still performs the best.
    Note that our method produces lower m2s than the reference method (achieved by MoSh++ over labeled markers), but it does not indicate which one is factually better than the other because 9.5mm offsets~\cite{mahmood2019amass} are expected between markers and a ``ground-truth'' body mesh.
    }
    \begin{tabular}{llccccccccc}
        \toprule
        \multirow{2}{*}{Method} & \multirow{2}{*}{Modality} & \multicolumn{3}{c}{Left Leg} & \multicolumn{3}{c}{Right Arm} & \multicolumn{3}{c}{Left Shoulder} 
        \\
        \cmidrule(lr){3-5}
        \cmidrule(lr){6-8}
        \cmidrule(lr){9-11}
        & & m2s$\downarrow$ & MPJPE$\downarrow$ & MPJVE$\downarrow$ & m2s$\downarrow$ & MPJPE$\downarrow$ & MPJVE$\downarrow$ & m2s$\downarrow$ & MPJPE$\downarrow$ & MPJVE$\downarrow$ \\
        \midrule
        VPoser & markers & 265.5 & 812.1 & 3137.0 & 165.2 & 652.8 & 2941.4 & 176.0 & 696.0 & 2913.2 \\
        HuMoR & markers & 241.6 & 721.9 & 2618.9 & 157.3 & 641.2 & 2555.3 & 180.2 & 632.2 & 2377.0 \\
        SOMA & markers & 106.6 & 678.1 & \textbf{509.5} & 53.0 & 451.5 & 625.8 & 53.6 & 716.8 & 610.0 \\
        \midrule
        HMR2.0+RR & markers+video & 42.0 & 301.3 & 751.0 & 25.4 & 280.6 & 506.2 & 32.1 & 413.2 & 610.5 \\
        VPoser+V & markers+video & 235.1 & 565.7 & 3413.7 & 210.4 & 620.7 & 3480.8 & 115.1 & 450.0 & 3003.5 \\
        HuMoR+V & markers+video & 309.3 & 627.1 & 2310.0 & 250.4 & 655.0 & 2405.4 & 148.2 & 458.9 & 1939.8 \\
        {\bf our method} & markers+video & \textbf{7.6} & \textbf{278.3} & 538.6 & \textbf{8.6} & \textbf{143.2} & \textbf{208.2} & \textbf{8.8} & \textbf{384.8} & \textbf{454.7} \\
        \midrule
        \cellcolor{gray!15}Reference & \cellcolor{gray!15}labeled markers & \cellcolor{gray!15}11.4 & \cellcolor{gray!15}0 & \cellcolor{gray!15}0 & \cellcolor{gray!15}10.2 & \cellcolor{gray!15}0 & \cellcolor{gray!15}0 & \cellcolor{gray!15}8.4 & \cellcolor{gray!15}0 & \cellcolor{gray!15}0 \\
        \bottomrule
    \end{tabular}
    \label{tab:comparisons_markers_partial}
\end{table*}
}

\subsection{Evaluation Protocols}
\label{ssec:protocol}

\subsubsection{Metrics}
We adopt multiple well-established metrics to comprehensively evaluate methods.
\begin{itemize}[noitemsep,topsep=0pt]
    \item \textbf{MPJPE} measures the mean per-joint position error. It is a common metric used to evaluate human pose reconstruction. Joint position errors are computed using the Euclidean distance between predicted and reference 3D joints.
    \item \textbf{MPJVE}  evaluates the mean estimated velocity error of each joint, computed on adjacent poses. This metric evaluates the temporal consistency of a motion. 
    \item \textbf{V2V}~\cite{SMPL-X:2019} computes the vertex-to-vertex error between the predicted and reference SMPL meshes. 
    It measures both body shape and pose.
    \item \textbf{m2s} measures the marker-to-surface distance. Intuitively, m2s measures the offset of the marker from the surface (skin) of the SMPL mesh. Real mocap markers have a marker offset, so there should be a small m2s even for the reference.
\end{itemize}
MJPVE is in millimeters per second and the others are in millimeters. For all the metrics, smaller values mean better performance.

\subsubsection{Datasets}
We use publicly available datasets that have hardware synchronized video and mocap markers.
Following the literature~\cite{rempe2021humor},
we downsample all datasets to 30Hz. 
We use MoSh++~\cite{mahmood2019amass} to generate reference data for both the CMU Kitchen Pilot dataset and the UMPM dataset.
To set up the UUO mocap, we remove marker labels in these datasets. Importantly, we do not use these datasets to train any models so that this simulates the unstructured setup.
Therefore, at best, methods can be trained on layouts in their training data but might not know layouts on the testing datasets.
\begin{itemize}[noitemsep,topsep=0pt]
    \item \textbf{CMU Kitchen} (pilot study)~\cite{de2009guide} is a challenging dataset due to self-occlusion and prevalence of stationary motions. Interestingly, individuals in this dataset wear backpacks and we remove the markers on the backpack as they are not near the appropriate body landmarks. It has 241 sequences and each is 15s long.
    
    \item \textbf{UMPM}~\cite{HICV11:UMPM} contains both markers and synchronized videos.
    As it uses it uses uncommon labels and marker placements, we create ground-truth label-vertex correspondences for this dataset.
    As we focus on single-person reconstruction, we use its single-person subset (p1) in our work.
    It has 12 sequences and each is 15s long.
    
    \item \textbf{MOYO}~\cite{tripathi20233d} contains many novel and difficult poses that are largely out-of-distribution compared to existing motion capture datasets.
    Importantly, we do not use this dataset to train models but only use its validation split for evaluation. This helps benchmark the generalization performance of different methods. 
    It has 171 sequences and each is 3s long.
\end{itemize}

\subsubsection{Compared methods}
We repurpose and compare existing mocap methods for UUO mocap. 
First, for the video-only method, we compare HMR2.0~\cite{goel2023humans}, a recent algorithm of monocular pose estimation. Because reconstruction from a monocular video contains ambiguities in global position, orientation, and scale, we evaluate its result with a rigid registration step to globally align the body w.r.t the mocap markers. We call this method HMR2.0+RR.
Moreover, we repurpose well-established marker-based methods for UUO mocap using appropriate modifications.
For example, SOMA~\cite{ghorbani2021soma} trains on diverse marker layouts and expects to generalize to unseen layouts.
We find that it has limited generalization in experiments; 
it trains on marker layouts that share marker labels with the CMU Kitchen and performs well on this dataset, but it performs significantly worse on other datasets (Table~\ref{tab:comparisons_markers_full}).
For SOMA, we run MoSh++ on the marker labels to solve for the SMPL parameters.
VPoser~\cite{SMPL-X:2019} and HuMoR~\cite{rempe2021humor} are not designed for UUO mocap but can be modified for it using appropriate constraints.
Importantly, both VPoser and HuMoR are sensitive to having strong initialization, so we initialize them with HMR 2.0's SMPL parameters. This results in modified versions that exploit both markers and video. We denote these as VPoser+V and HuMoR+V.
Finally, we use MoSh++~\cite{mahmood2019amass} as the reference achieved by using \emph{labeled} markers.

\subsection{Comparisons}

Tables~\ref{tab:comparisons_markers_full} and \ref{tab:comparisons_markers_partial} benchmark methods for full- and partial-body reconstruction, respectively.
We summarize salient conclusions here.
First, our method outperforms the compared approaches by a large margin on all datasets for both full- and partial-body mocap.
Second, 
SOMA, a method that involves labeling markers from a database of diverse full-body marker layouts, performs competitively against our method for full-body mocap (Table~\ref{tab:comparisons_markers_full}) but performs poorly for partial-body mocap.
This demonstrates the challenge of mocap w.r.t unstructured markers, e.g. for partial-body reconstruction.
Third, VPoser and HuMoR, 
originally designed for mocap using labeled markers, perform poorly with unlabeled markers. Assisted by monocular video, they (VPoser+V and HuMoR+V) generally achieve significant improvement for full-body reconstruction (Table~\ref{tab:comparisons_markers_full}).
However, for partial-body reconstruction,
all other methods perform significantly worse than ours (Table~\ref{tab:comparisons_markers_partial}), further confirming the difficulty of solving mocap using unstructured markers. Our method still outperforms others for partial-body mocap.

We provide visualizations in Figs.~\ref{fig:qualitative_moyo_val}, \ref{fig:qualitative_umpm}, and \ref{fig:qualitative_cmu_kitchen} to qualitatively compare different methods. We also attach video demos in the supplementary material.
Visual results show that VPoser+V and HuMoR+V both struggle to find correct root alignment during optimization, even with a video prior. 
Both additionally struggle to generate poses that match the markers well for the UMPM dataset, often resulting in more static poses. In addition, these two methods tend to produce far more jittery motions. Recall that SOMA is trained on a few different marker layouts, it chooses marker labels a superset of positions of these layouts; as UMPM has an unusual marker layout, SOMA does not perform well on it. On a 15-second UMPM sequence data, our method takes ~12min (plus ~6min for video processing), while SOMA (GPU) and MoSh++(CPU) takes ~20min (i5-13600k/Titan RTX).

\subsection{Ablation Study}
We perform two main ablations studies to demonstrate the effectiveness of different parts of our pipeline.

\begin{table}
    \centering
    \small
    \caption{\small Ablation for multi-hypothesis testing (MHT)  for root orientation (Sec.~\ref{ssec:mocap-solving}, Stage 1) on the UMPM dataset~\cite{HICV11:UMPM}. 
    ``- MHT'' means that we remove MHT and only optimize the initial orientation. Using multiple initial rotations avoids local minima during mesh alignment and significantly improves mocap performance.
    }
    \label{tab:ablations_system}
    \begin{tabular}{lccccc}
        \toprule
        Method & m2s$\downarrow$ & MPJPE$\downarrow$ & MPJVE$\downarrow$ & V2V$\downarrow$ \\
        \midrule
        our method (that uses $|\mathcal{B}|$=4) & \textbf{11.0} & \textbf{60.8} & \textbf{81.5} & \textbf{62.6} \\
        \ \ \ \ - MHT (i.e., $|\mathcal{B}|$=1) & 26.6 & 297.0 & 421.5 & 395.0 \\
        \bottomrule
    \end{tabular}
\end{table}

\begin{table}
    \centering
    \small
    \caption{\small Stage ablations on the UMPM~\cite{HICV11:UMPM} dataset for the full-body reconstruction task (Sec.~\ref{ssec:mocap-solving}). 
    We progressively evaluate the result after each stage. Note that Stage 3 finds marker-vertex correspondence and does not involve optimization.
    Results show that the stage-wise optimization clearly leads to better mocap performance.
    }
    \label{tab:ablations_stage}
    \begin{tabular}{lccccc}
        \toprule
        Stage & m2s$\downarrow$ & MPJPE$\downarrow$ & MPJVE$\downarrow$ & V2V$\downarrow$ \\
        \midrule
        \ \ Marker-part matching & 69.8 & 240.4 & 651.5 & 283.4 \\
        \ \ Stage 2: pose fitting & 15.7 & 88.1 & 620.2 & 88.3 \\
        \ \ Stage 4: inverse kinematics & 11.6 & 62.1 & 89.3 & 63.6 \\
        \ \ Stage 5: solver refinement & \textbf{11.0} & \textbf{60.8} & \textbf{81.5} & \textbf{62.6} \\
        \bottomrule    
    \end{tabular}
\end{table}

\subsubsection{System design}
We highlight some of the key design choices in Table.~\ref{tab:ablations_system}. We show that having multi-hypothesis testing (MHT) is critical for correct alignment for solving. Our method that uses $\mathcal{B}=\{0^{\circ}, 90^{\circ}, 180^{\circ}, 270^{\circ}\}$ performs substantially better than solving for rotation with a single initial starting angle (i.e., $\mathcal{B}=\{0^{\circ}\}$).

\subsubsection{Stage ablations}
Table~\ref{tab:ablations_stage} measures the errors produced after each stage in the third module Mocap Solver (Sec.~\ref{ssec:mocap-solving}).
We use the best angle as determined after MHT for root orientation to evaluate stages 2 and 4. Stage 2, the pose fitting stage, is dominated by a Chamfer distance loss and finds best fit on a per-frame basis.
Unfortunately, this stage adds considerable jitter, which is reflected by the high MPJVE error. 
The Stage 4 inverse kinematics effectively removes this jitter because the marker-surface correspondence is locked during the optimization process. 
Finally, Stage 5 solver refinement helps to mitigate the effect of incorrect marker locations.

\section{Limitations}
\label{ssec:limitations}
\begin{figure*}
    \centering
    \begin{subfigure}{0.45\textwidth}
        \includegraphics[width=\linewidth]{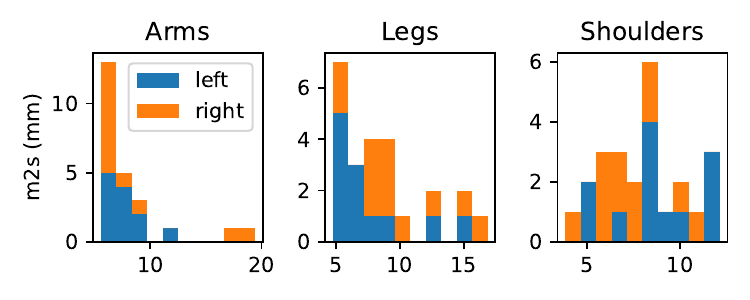}
    \end{subfigure}
    \begin{subfigure}{0.45\linewidth}
        \includegraphics[width=\linewidth]{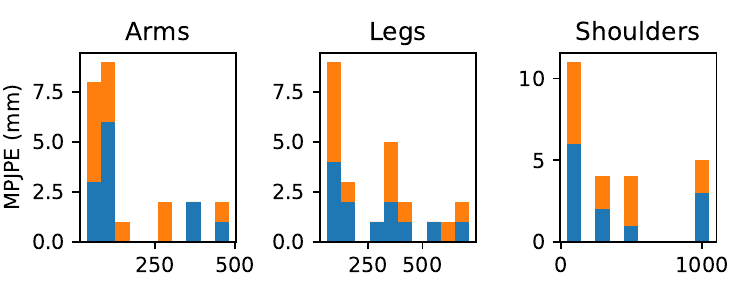}
    \end{subfigure}
    \caption{Frequency of errors (left: m2s, right: MPJPE) for individual sequences for partial-body reconstruction. One limitation of our algorithm is that matching the wrong part can lead to higher errors.}
    \Description{Frequency of errors (left: m2s, right: MPJPE) for individual sequences for partial-body reconstruction. One limitation of our algorithm is that matching the wrong part can lead to higher errors.}
    \label{fig:part_error_distribution}
\end{figure*}

Our approach sometimes struggles with aligning the correct part to markers, especially when they lack certain identifiable characteristics (Fig.~\ref{fig:part_error_distribution}). For example, if an actor only has markers on the left leg and jumps with both legs, then the wrong leg could be aligned to the markers. In practice, this may be less of an issue because the use of unilateral partial-body reconstruction is often for isolation movements in biomechanical analysis. Another issue is that very sparse layouts (e.g., UMPM) can be labeled incorrectly due poor coverage. Incorporating physical scene constraints could help (e.g.\  the floor) improve reconstruction.

\section{Conclusion}
\label{ssec:conclusion}
We motivate the problem of mocap with unstructured unlabeled optical (UUO) markers.
We propose to exploit a monocular video, captured alongside markers, to estimate a human body prior. 
Concretely, we extract an initial SMPL model from the video, and use it to optimize human body shape, pose, global translation, and rotation by fitting the SMPL model to the UUO markers.
We introduce a pipeline of optimization techniques and show its superior performance over prior art on three benchmark UUO mocap datasets for both full-body and partial-body mocap.

\begin{acks}
Authors thank Dr. Lingjie Liu for insightful discussions.
Shu Kong is partially supported by the University of Macau (SRG2023-00044-FST).
The CMU Kitchen data used in this paper was obtained from kitchen.cs.cmu.edu and the data collection was funded in part by the National Science Foundation under Grant No. EEEC-0540865.
\end{acks}

\bibliographystyle{ACM-Reference-Format}
\bibliography{references}

\newcommand{\sswidth}{0.12\textwidth}

\begin{figure*}
    \centering
    \begin{tabular}{cccc}
        HMR 2.0+RR & SOMA+Mosh++ & Ours & Reference from MOYO \\
        
        \includegraphics[width=\sswidth]{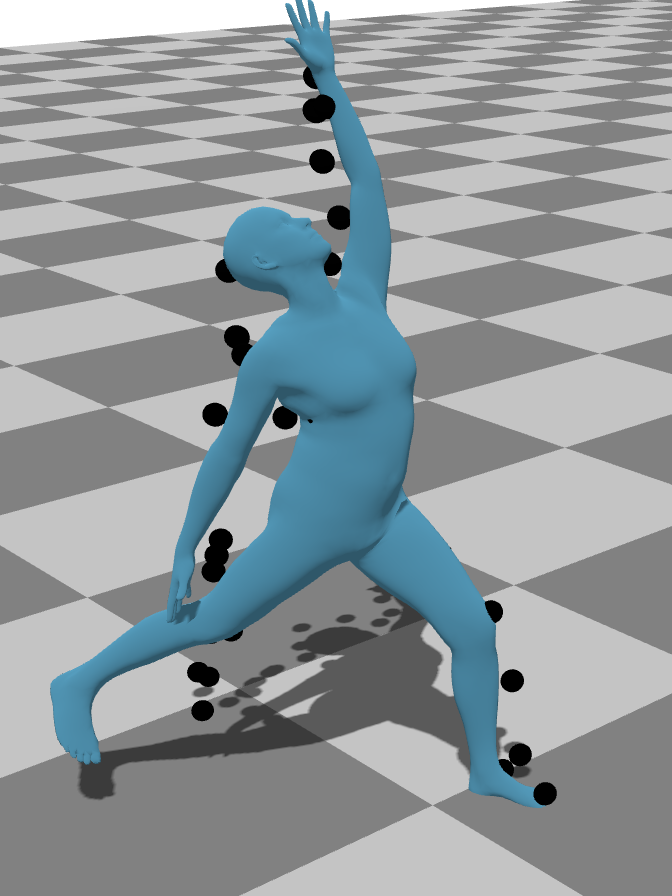} &
        \includegraphics[width=\sswidth]{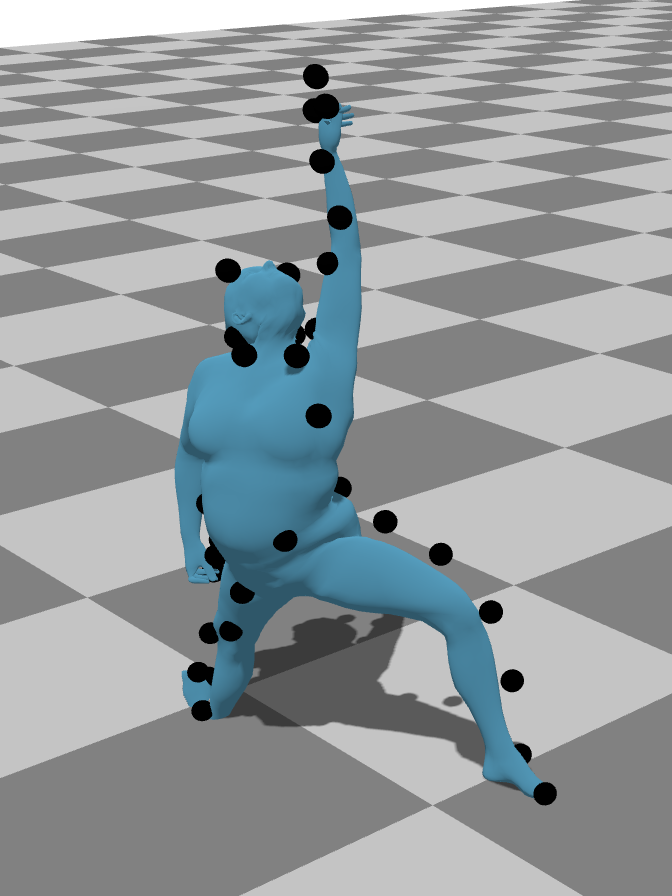} &
        \includegraphics[width=\sswidth]{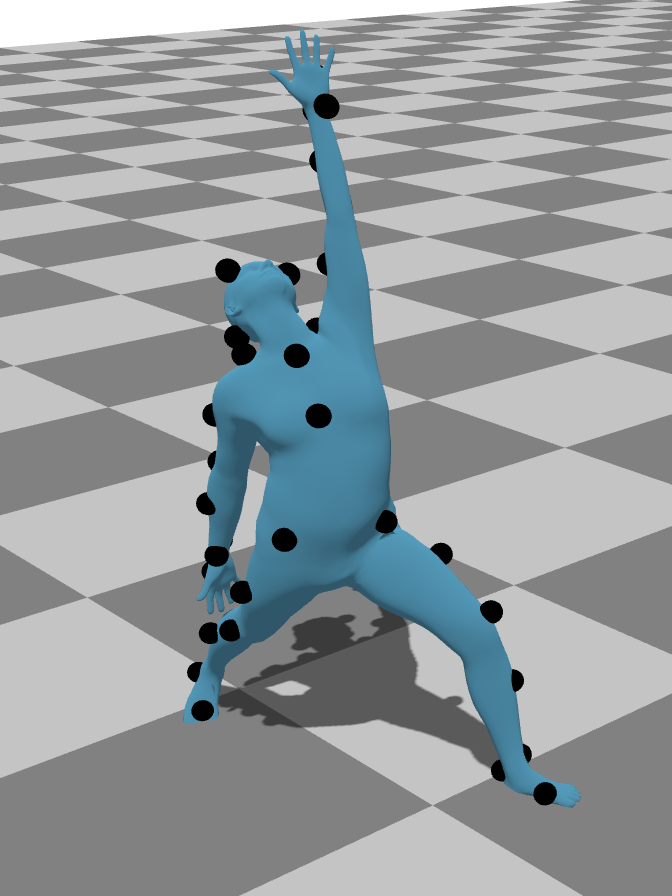} &
        \includegraphics[width=\sswidth]{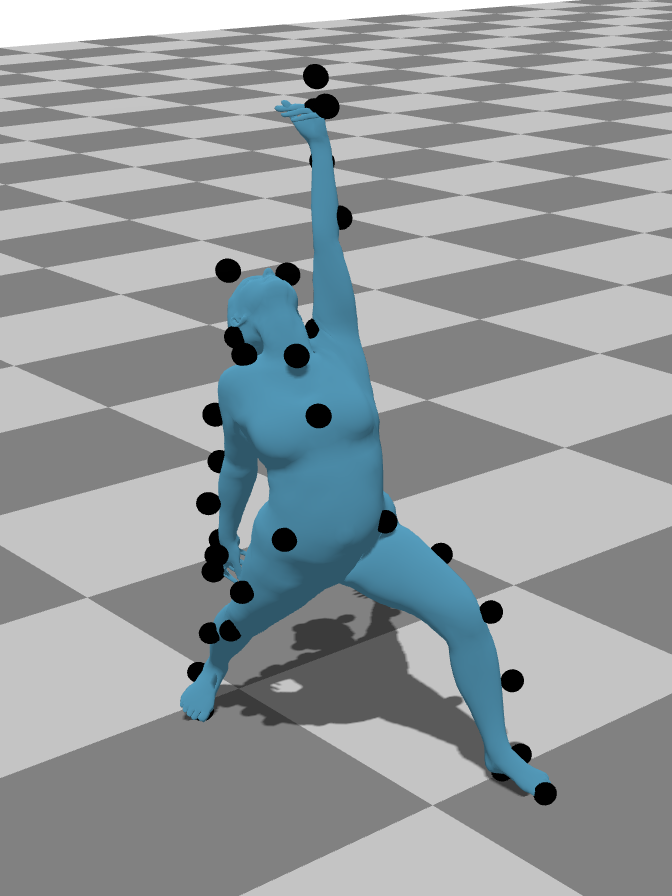} \\
        
        \includegraphics[width=\sswidth]{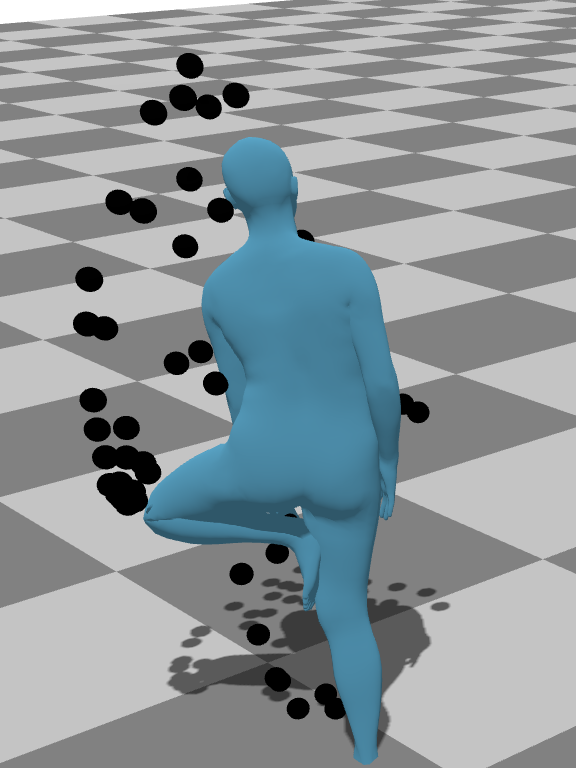} &
        \includegraphics[width=\sswidth]{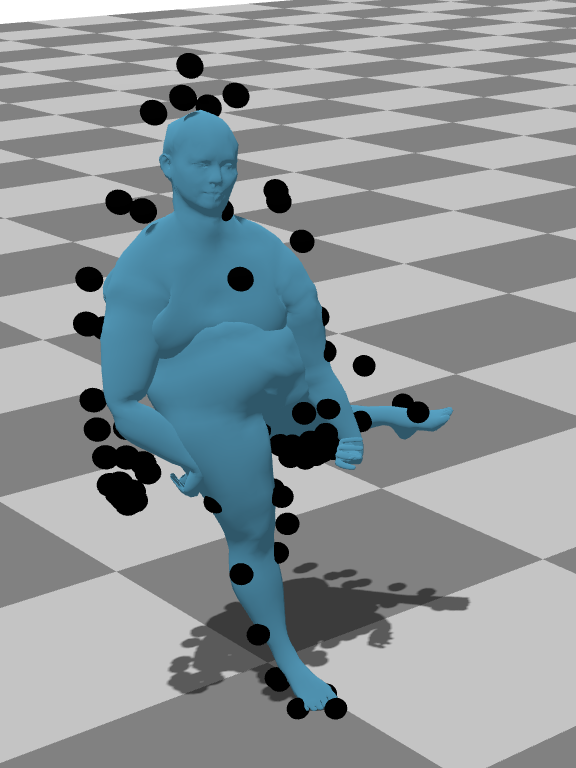} &
        \includegraphics[width=\sswidth]{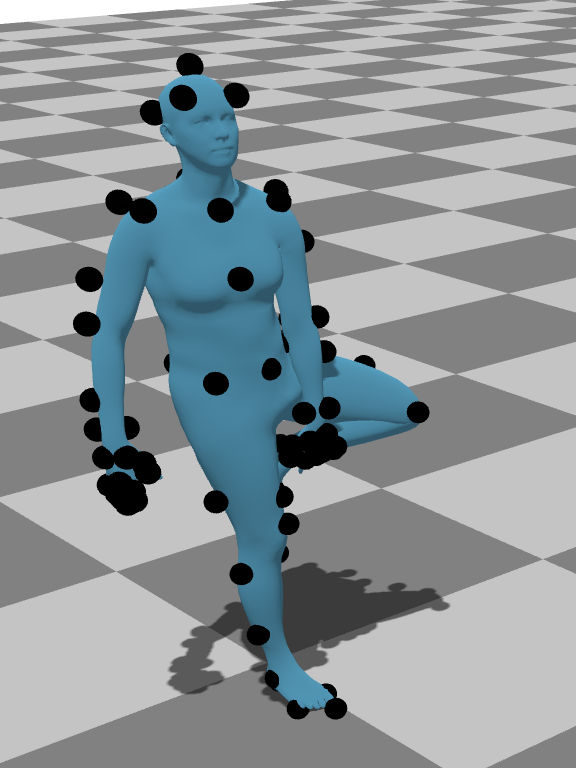} &
        \includegraphics[width=\sswidth]{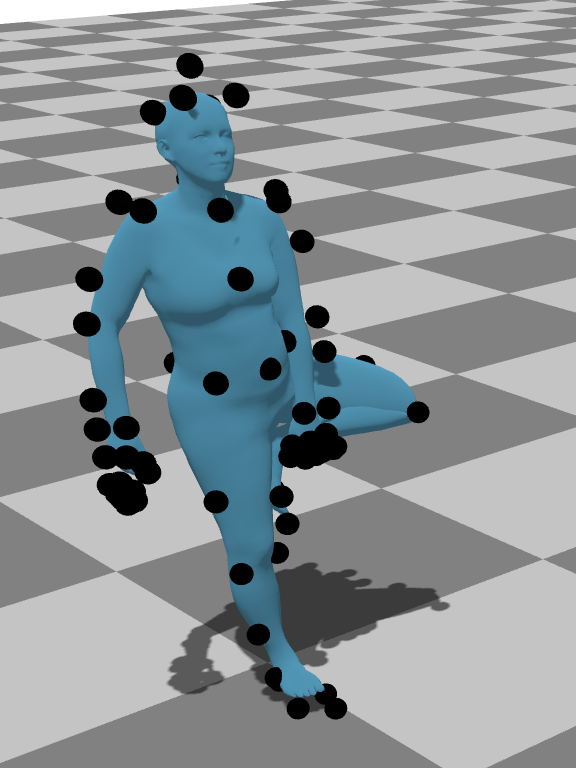} \\

        \includegraphics[width=\sswidth]{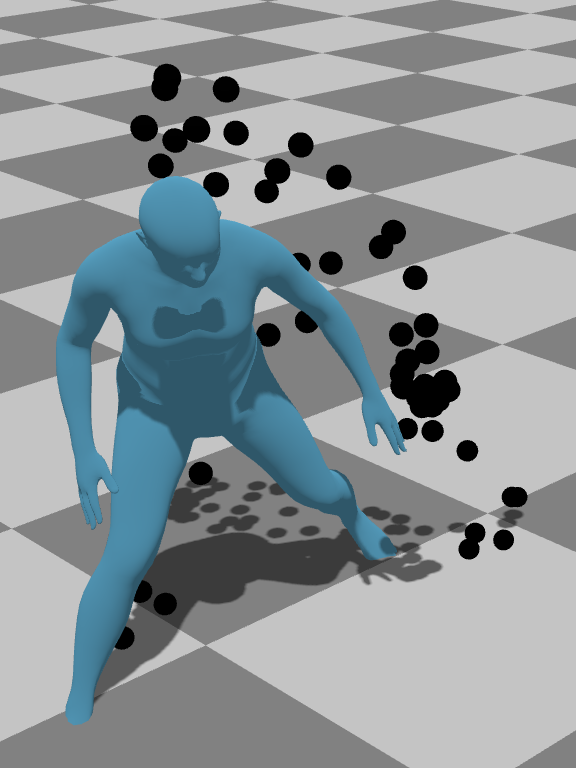} &
        \includegraphics[width=\sswidth]{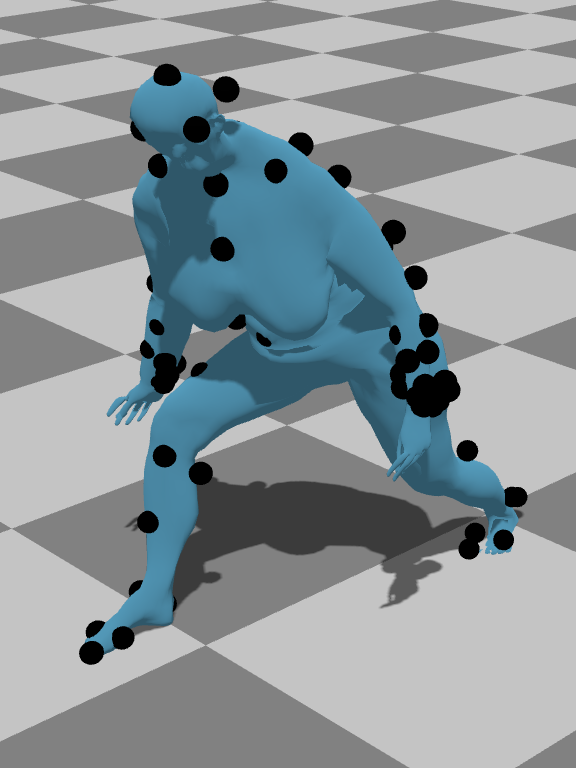} &
        \includegraphics[width=\sswidth]{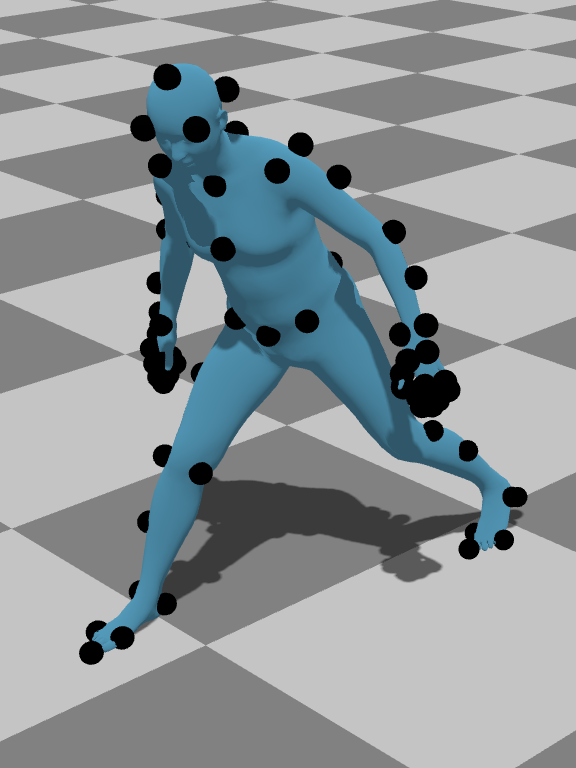} &
        \includegraphics[width=\sswidth]{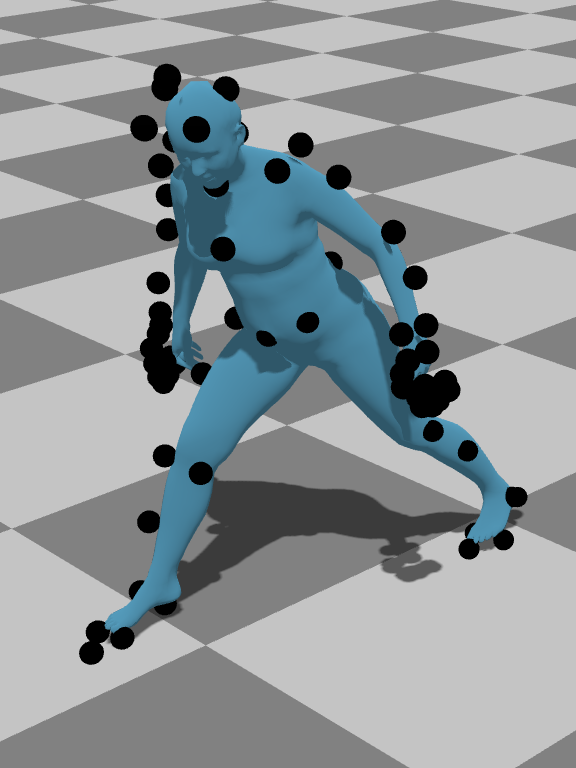} \\
    \end{tabular}
    \caption{Qualitative results for the validation split of the MOYO dataset~\cite{tripathi20233d}. This dataset is challenging that has unique and difficult poses. Furthermore, markers are densely packed, which can present ambiguity for labeling. SOMA struggles to accurately label the markers, resulting in poor quality reconstruction.
    Our method produces better visual results.
    }
    \Description{Qualitative results for the validation split of the MOYO dataset~\cite{tripathi20233d}. This dataset is challenging that has unique and difficult poses. Furthermore, markers are densely packed, which can present ambiguity for labeling. SOMA struggles to accurately label the markers, resulting in poor quality reconstruction.
    Our method produces better visual results.
    }
    \label{fig:qualitative_moyo_val}
\end{figure*}

\begin{figure*}
    \centering
    \begin{tabular}{cccc}
        HMR 2.0+RR & SOMA+Mosh++ & Ours & Reference \\

        \includegraphics[width=\sswidth]{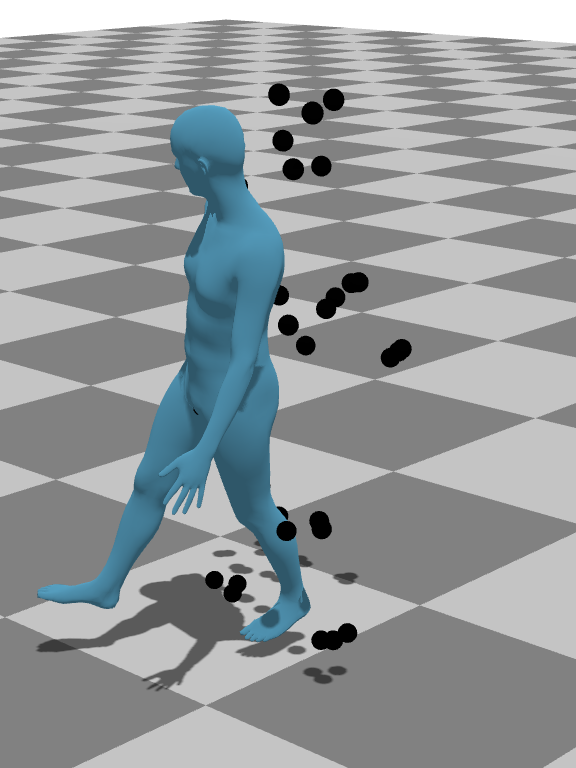} &
        \includegraphics[width=\sswidth]{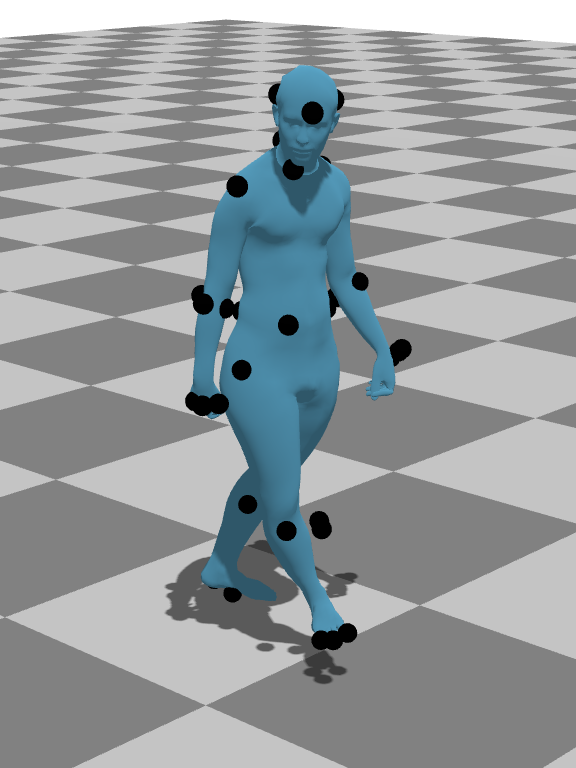} &
        \includegraphics[width=\sswidth]{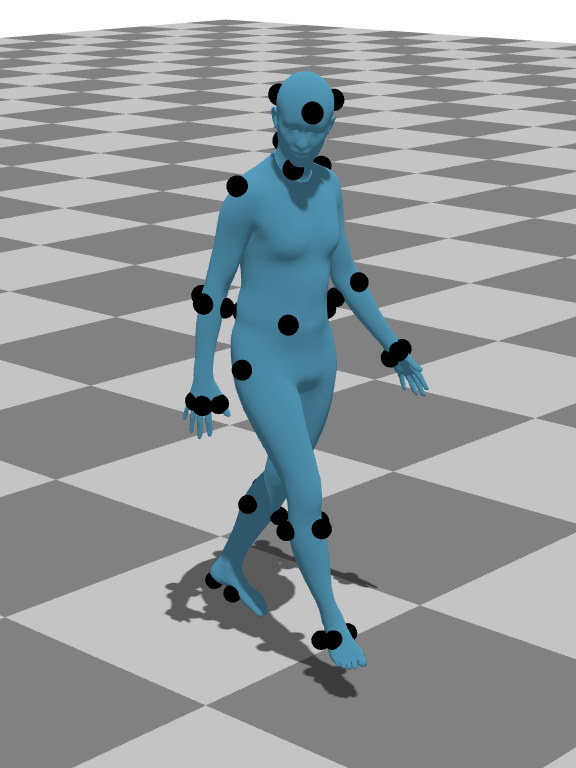} &
        \includegraphics[width=\sswidth]{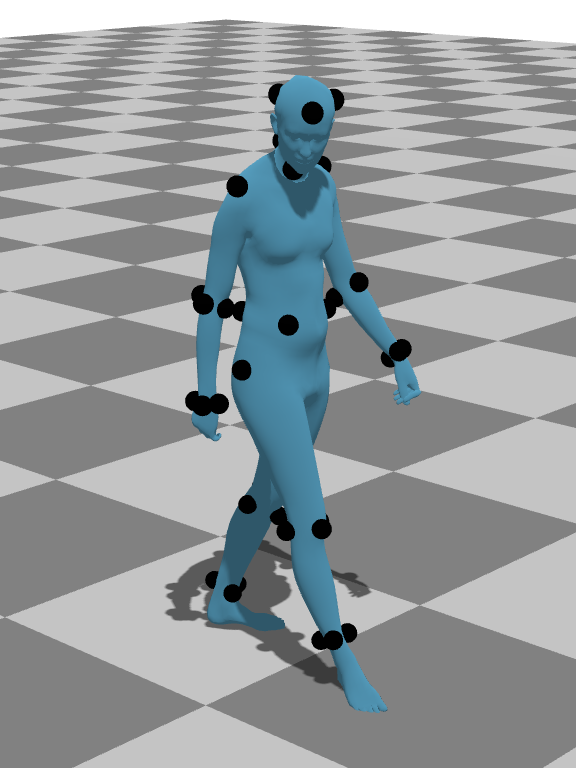} \\

        \includegraphics[width=\sswidth]{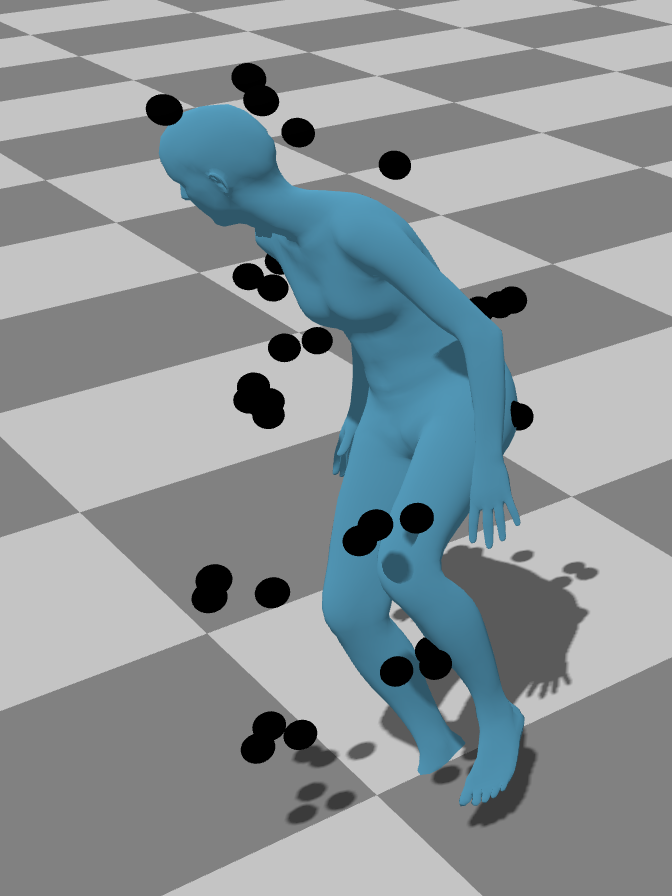} &
        \includegraphics[width=\sswidth]{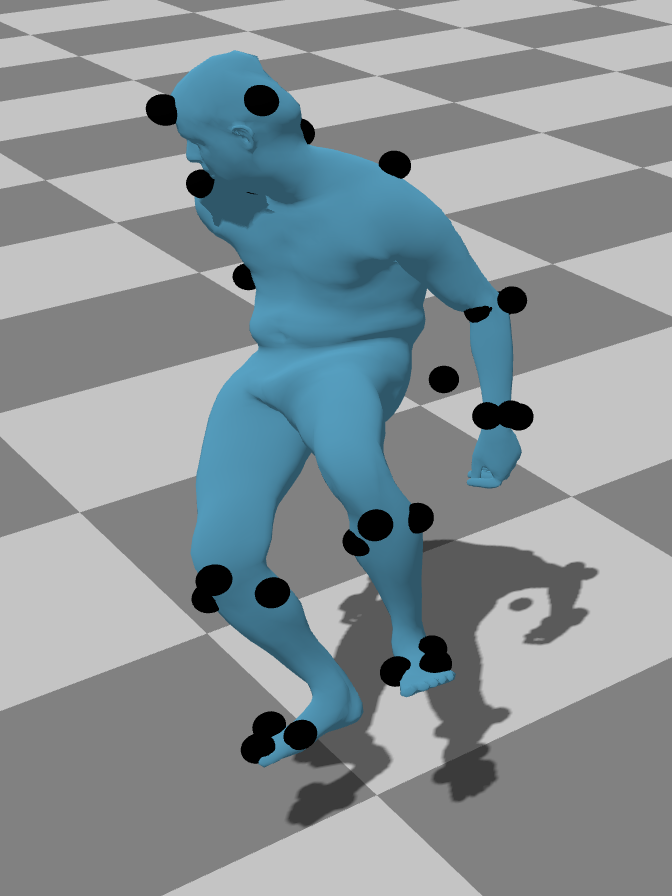} &
        \includegraphics[width=\sswidth]{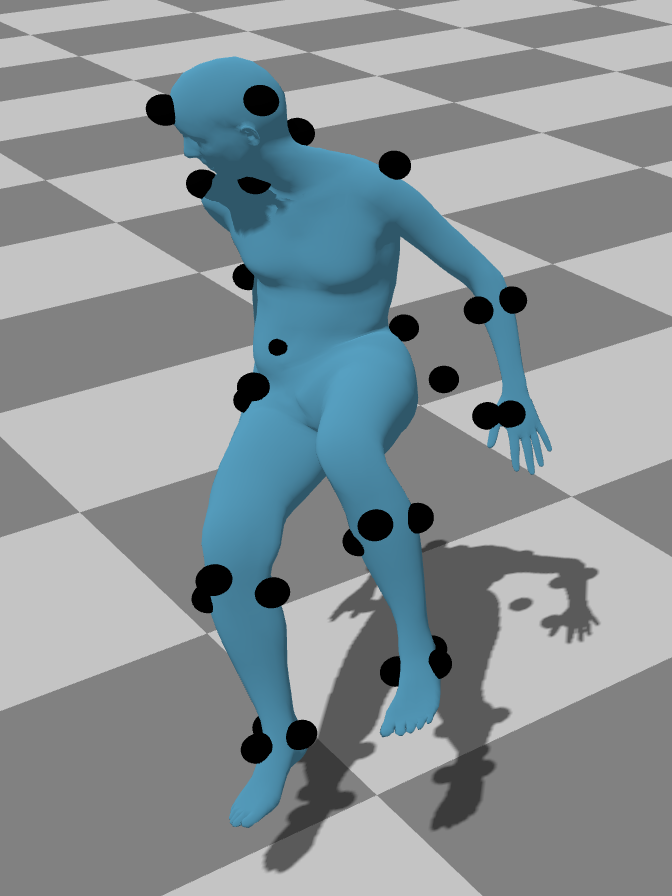} &
        \includegraphics[width=\sswidth]{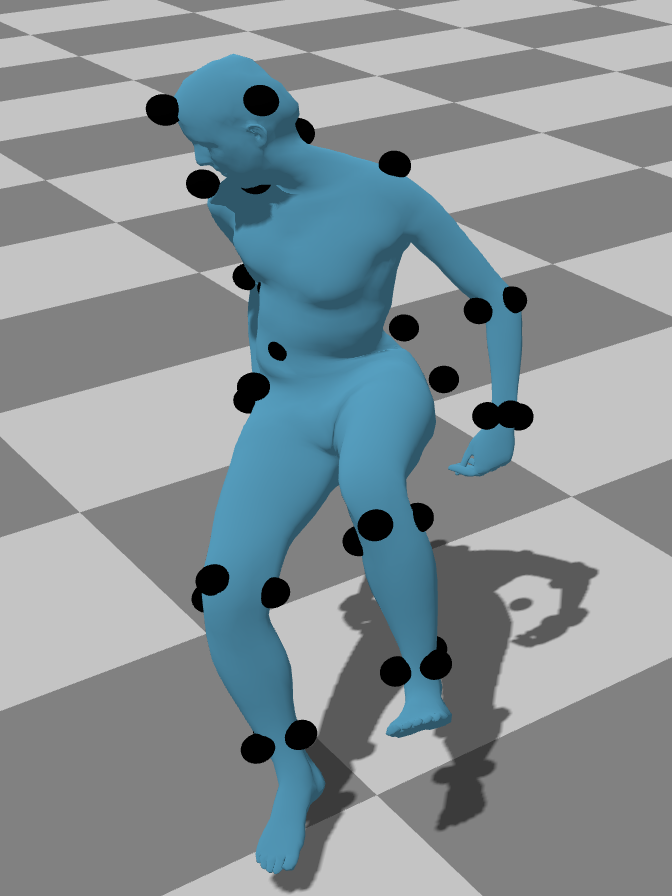} \\
    
    \end{tabular}
    \caption{Qualitative results for the validation split of the UMPM dataset~\cite{HICV11:UMPM}. HMR2.0+RR contains alignment issues, and SOMA produces an incorrect joint position at the right knee. In contrast, our method produces better visual results.}
    \Description{Qualitative results for the validation split of the UMPM dataset~\cite{HICV11:UMPM}. HMR2.0+RR contains alignment issues, and SOMA produces an incorrect joint position at the right knee. In contrast, our method produces better visual results.}
    \label{fig:qualitative_umpm}
\end{figure*}

\begin{figure*}
    \centering
    \begin{tabular}{cccc}
        HMR 2.0+RR & SOMA+Mosh++ & Ours & Reference \\
        
        \includegraphics[width=\sswidth]{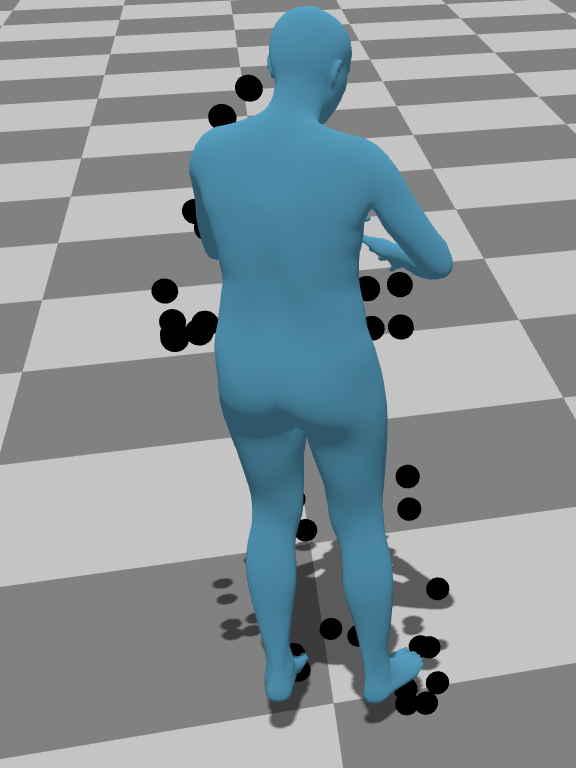} &
        \includegraphics[width=\sswidth]{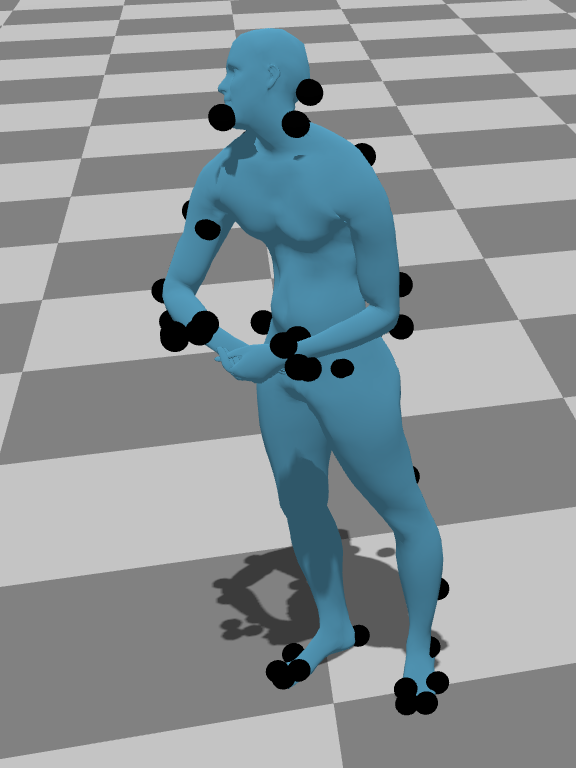} &
        \includegraphics[width=\sswidth]{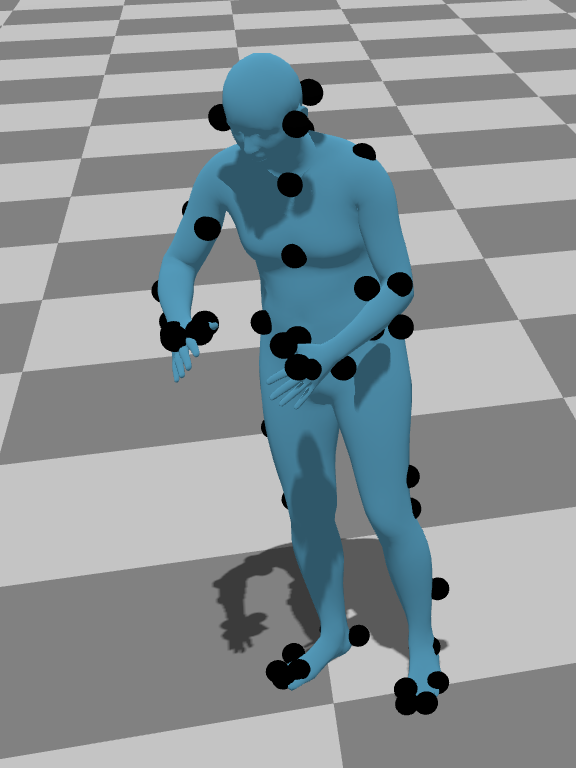} &
        \includegraphics[width=\sswidth]{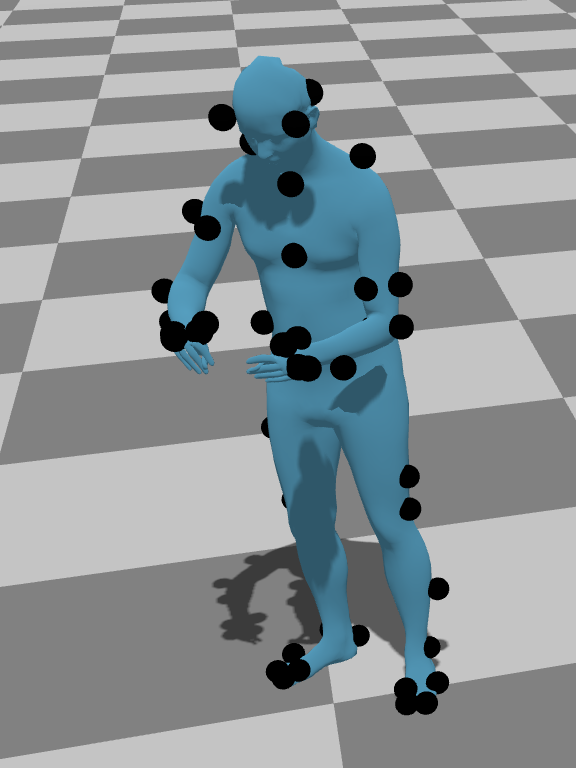} \\
        
        \includegraphics[width=\sswidth]{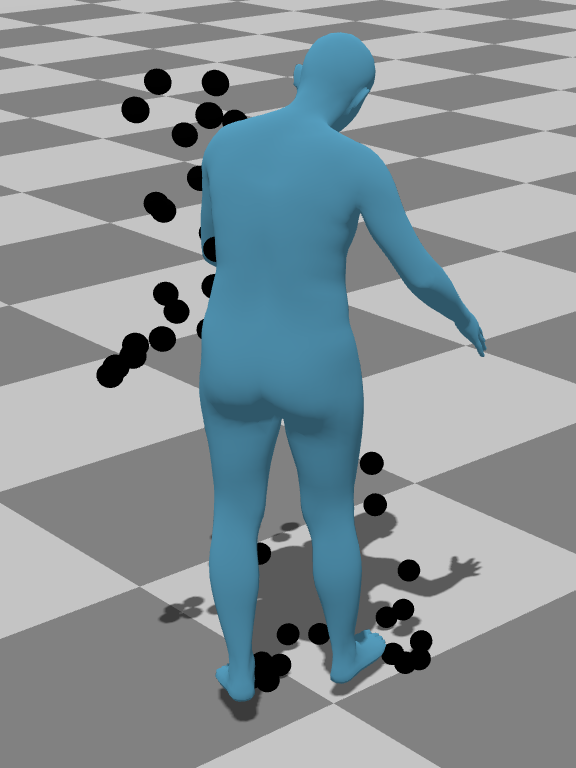} &
        \includegraphics[width=\sswidth]{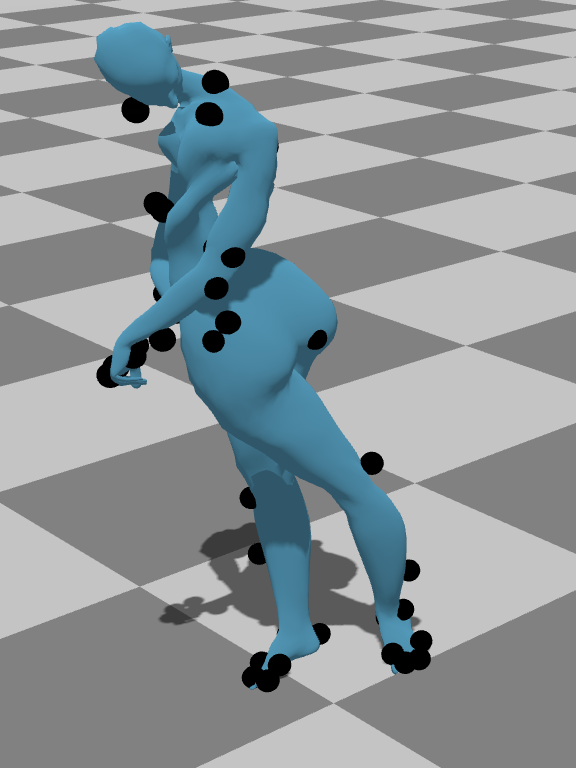} &
        \includegraphics[width=\sswidth]{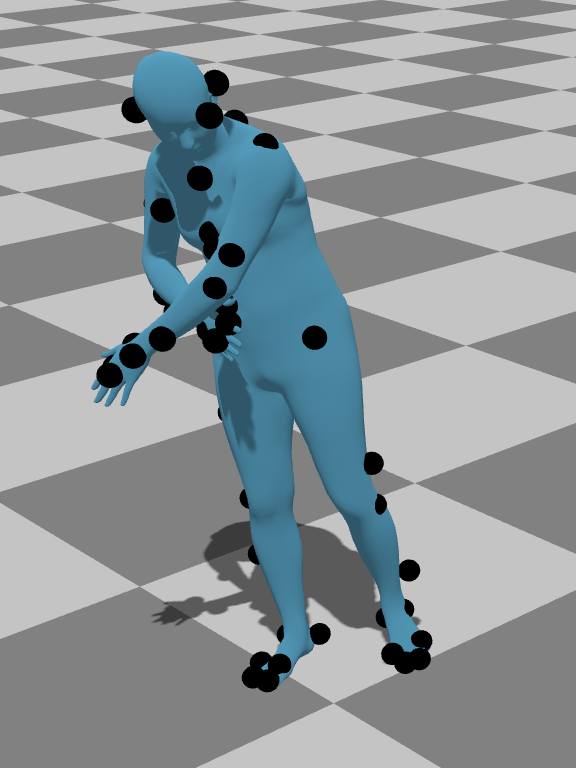} &
        \includegraphics[width=\sswidth]{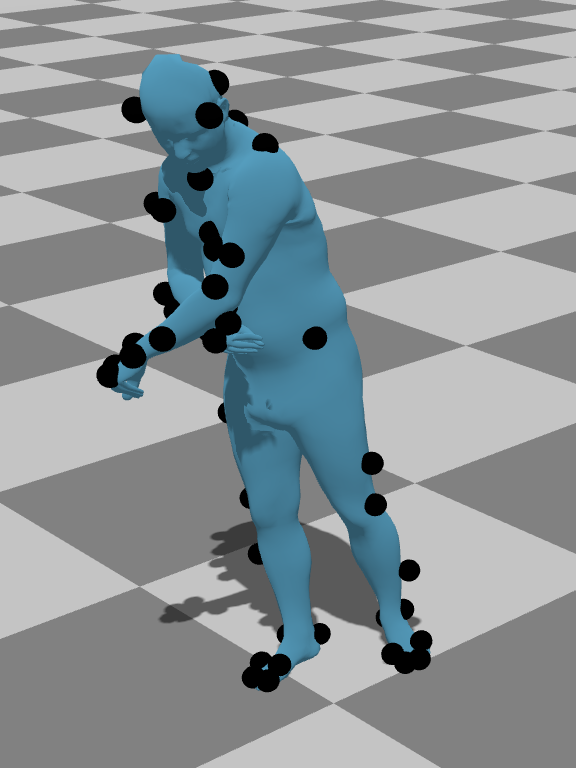} \\

    \end{tabular}
    \caption{Qualitative results for the validation split of the CMU Kitchen dataset~\cite{de2009guide}. Our approach does aligns better to the markers compared to HMR 2.0+RR and produces a closer body shape and poser to the reference compared to SOMA.}
    \Description{Qualitative results for the validation split of the CMU Kitchen dataset~\cite{de2009guide}. Our approach does aligns better to the markers compared to HMR 2.0+RR and produces a closer body shape and poser to the reference compared to SOMA.}
    \label{fig:qualitative_cmu_kitchen}
\end{figure*}

\begin{figure*}
    \centering
    \begin{tabular}{cccc}
        HMR 2.0+RR & SOMA+Mosh++ & Ours & Reference \\
        \includegraphics[width=\sswidth]{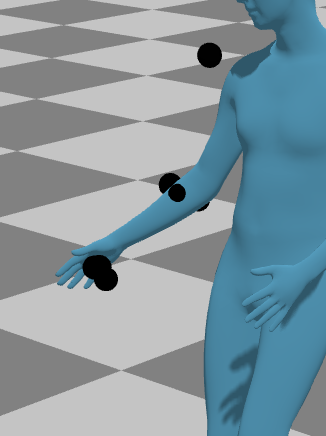} &
        \includegraphics[width=\sswidth]{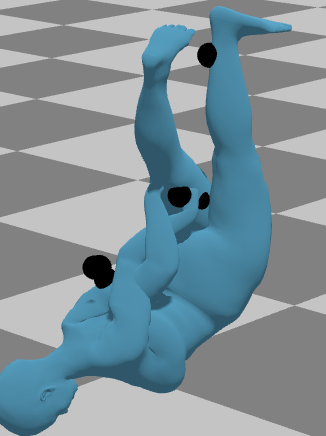} &
        \includegraphics[width=\sswidth]{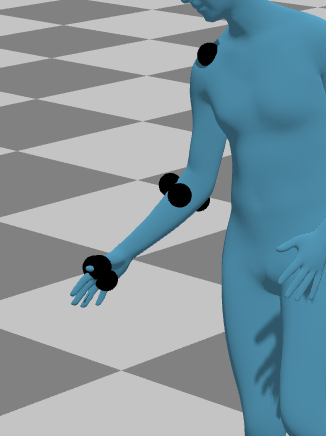} &
        \includegraphics[width=\sswidth]{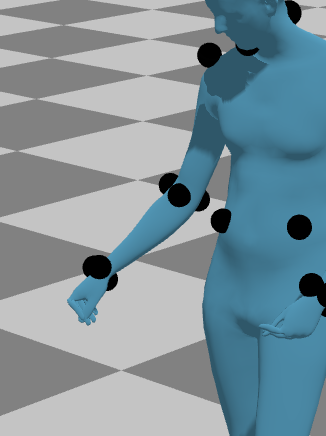} \\
        
        \includegraphics[width=\sswidth]{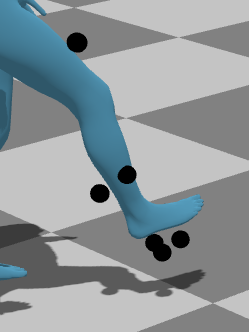} &
        \includegraphics[width=\sswidth]{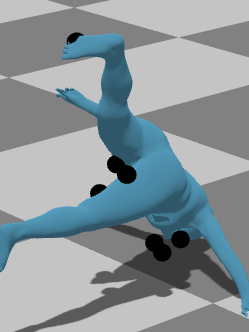} &
        \includegraphics[width=\sswidth]{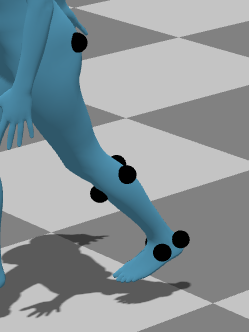} &
        \includegraphics[width=\sswidth]{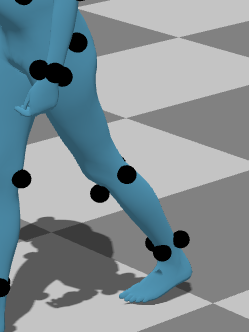} \\
    \end{tabular}
    \caption{Partial-body reconstruction for the UMPM dataset~\cite{HICV11:UMPM} for right arm (top row) and left leg (bottom row).
    SOMA is unable to handle partial body reconstruction;
    HMR 2.0+RR aligns the correct part due to using our part localization but has noticeable gaps between the markers and the surface and incorrect alignment.
    Our method produces better visual results.
    }
    \Description{Partial-body reconstruction for the UMPM dataset~\cite{HICV11:UMPM} for right arm (top row) and left leg (bottom row).
    SOMA is unable to handle partial body reconstruction;
    HMR 2.0+RR aligns the correct part due to using our part localization but has noticeable gaps between the markers and the surface and incorrect alignment.
    Our method produces better visual results.
    }
    \label{fig:part_based reconstruction}
\end{figure*}

\appendix

\appendix

\section{Parts}
For part evaluation, we use the part definitions in Table~\ref{tab:part_definitions}. Each part consists of multiple bones (i.e., SMPL "joints").

\begin{table*}[ht]
    \centering
    \caption{\small Part definitions with corresponding SMPL bone names}
    \label{tab:part_definitions}
    \begin{tabular}{ll}
         \toprule
         Part name & Joints \\
         \midrule
         Left arm & left\_shoulder, left\_elbow, left\_wrist \\
         Left leg & left\_hip, left\_knee, left\_ankle, left\_foot \\
         Left shoulder & spine3, left\_collar, left\_shoulder, left\_shoulder, left\_elbow \\
         Right arm & right\_shoulder, right\_elbow, right\_wrist \\
         Right leg & right\_hip, right\_knee, right\_ankle, right\_foot \\
         Right shoulder & spine3, right\_collar, right\_shoulder, right\_shoulder, right\_elbow \\
         \bottomrule
    \end{tabular}
\end{table*}

\section{Comparisons}
In this section, we discuss experimental setups of comparison techniques.

\subsection{Reference}
\subsubsection{UMPM and CMU Kitchen}. Our reference data is computed by using labeled markers with MoSh++~\cite{mahmood2019amass}. However, one problem is that there is variation with the marker placements. While MoSh++ can handle some variation in placement, too much variation can cause error in reconstruction. In our dataset, we observe some errors in fitting markers to joint positions.

For the UMPM~\cite{HICV11:UMPM} dataset, we manually create the marker-label correspondence between UMPM marker labels and SMPL-X~\cite{SMPL-X:2019} vertices. During manual labeling, we cross referenced various videos of the actors and the manual~\cite{HICV11:UMPM} to determine marker placement. Markers are often placed along a band around limbs, but the orientation of this band can vary and cause some errors in reconstruction. Thus, we report m2s as well as other traditional pose metrics.

The CMU Kitchen Pilot~\cite{de2009guide} dataset uses common marker labels, so we use the labels provided in the MoSh++ source code. One problem with this dataset is that the actors wear a backpack with 7 markers (LBWT, NEWLBAC, NEWRBAC, RBAC, RBWT, T10, T8), on the exterior of the backpack. These marker labels traditionally correspond to markers placed on the back, but the backpack adds offsets that distort the body shape considerably. Thus, we test both with and without these markers. While this dataset used hardware synchronization, we found that the source files are not synchronized. Each clip has a synchronization event in which the actor turns on and off a light bulb at the start and end of each trial. We manually synchronize the video and mocap data using these events. Furthermore, we found that the video data is closer to 29.97Hz while the source mocap data is at 120Hz (which we downsample to 30Hz). The discrepancy between frequencies becomes an issue for longer video sequences. To account for this, we insert a duplicate video frame to change the frequency to 30Hz.

For all of the datasets, we use 12 evenly-spaced frames (starting with the first frame and ending with the last frame) to perform the first stage in MoSh++~\cite{mahmood2019amass}.

\subsubsection{MOYO}. For the MOYO~\cite{tripathi20233d} dataset, we use the SMPL-X models provided by the authors.

\subsubsection{Conversion}. For all three datasets, we need to convert from the SMPL-X model to the neutral SMPL body model for evaluation. More specifically, we convert SMPL-X models to neutral SMPL models via the official conversions tools (https://github.com/vchoutas/smplx).

\subsection{HuMoR}
For motion capture solving~\cite{rempe2021humor}, HuMoR requires marker labels in the form of vertex correspondences. However, they also test on RGB-D datasets. In this case, they use a Chamfer distance loss. We apply a single-directional Chamfer distance loss without robust weighting as we found this to help with reconstruction. Additionally, as HuMoR is more efficient for optimizing smaller time window, we adopt their sequence splitting and merging method wherein we split the sequence into overlapping sequences of 2 seconds. We found it necessary to tune some of the parameters used in their approach:
\begin{verbatim}
--data-fps 30

--prox-batch-size 8  # 2 for MOYO, 8 for UMPM and CMU Kitchen

--prox-seq-len 60
--prox-overlap-len 1
--point3d-weight 100000.0 100000.0 100000.0
--pose-prior-weight 0.01 0.01 0.0
--shape-prior-weight 0.1 0.1 0.1

--joint3d-smooth-weight 0.1 0.1 0.0

--motion-prior-weight 0.0 0.0 1e-5
--motion-optim-shape

--init-motion-prior-weight 0.0 0.0 0.0

--joint-consistency-weight 0.0 0.0 100.0
--bone-length-weight 0.0 0.0 2000.0

--contact-vel-weight 0.0 0.0 100.0
--contact-height-weight 0.0 0.0 10.0

--floor-reg-weight 0.0 0.0 1.0

--lr 1.0
--num-iters 30 70 70

--stage3-tune-init-num-frames 15
--stage3-tune-init-freeze-start 30
--stage3-tune-init-freeze-end 55
\end{verbatim}

\subsection{SOMA}
SOMA~\cite{ghorbani2021soma} labels markers from generally structured marker layouts. We compare their SuperSet model, which is trained to classify 89 common marker labels. The idea behind the SuperSet is that there are many common marker layouts that share selected keypoints on the body for humans. Note that the SuperSet model is generally less accurate than fine-tuned models per layout, but it is necessary to use this model when the marker layout is unknown. SOMA provides these discrete marker labels, and then MoSh++ uses these labels to compute the SMPL-X parameters for the human. MoSh++ uses two stages of optimization. The goal of the first stage is to estimate the body shape and the marker locations on the surface. The goal of the second stage is to estimate the pose. We repeat both stages for every sequence. Furthermore, MoSh++ requires 12 representative frames for the first stage. For this we simply select 12 frames uniformly spaced across the entire sequence.

\section{Alignment of monocular video and mocap markers}
HMR 2.0~\cite{goel2023humans} provides SMPL and camera parameters. However, it does not provide accurate root translation, as root translation in world-space is a difficult problem to solve for monocular video~\cite{yuan2022glamr, ye2023decoupling, shin2023wham}. The root orientation from HMR 2.0 does not necessarily align with the root orientation of the mocap markers. The difference is mostly due to a single yaw-rotational offset between the root orientations of HMR and mocap markers. Optimizing for root orientation is prone to local minima, particularly with respect to front and back of the SMPL mesh.

\section{Data Preprocessing}
The motion capture data generally is high quality in all three datasets. However, we needed to handle some edge cases in preprocessing for a small number of sequences. During some frames, markers could reset to the origin, which may have been caused by tracking errors. These markers are masked for the problematic frames during the optimization process. HMR 2.0~\cite{goel2023humans} with PHALP~\cite{rajasegaran2022tracking} sometimes drops tracking for some frames. If these frames are at the beginning or end of the sequences, we use the closest known SMPL parameters. If the frames are in the middle of the sequences, we linearly interpolate $\beta$ and $\Gamma$ and perform spherical linear interpolation~\cite{bregier2021deepregression} $\Phi$ and $\Theta$. However, we mask out these frames with our method when finding the marker-vertices correspondences.

\subsection{Video Processing}
All three datasets have multiple cameras with different labels. Because we only evaluate with monocular vision, we select one camera for each dataset. Furthermore, we down-sample each dataset to reduce video processing time. The video properties and resolutions are shown in Table~\ref{tab:cam_labels}.

\begin{table}[ht]
    \centering
    \caption{\small We list the configurations for the community to reproduce results and fairly benchmark results.}
    \label{tab:cam_labels}
    \begin{tabular}{@{}lccc@{}}
        \toprule
         & UMPM & CMU Kitchen & MOYO \\
        \midrule
        Camera name & l & 7151062 & YOGI\_Cam\_06 \\
        Resolution & $644\times486$ & $1024\times768$ & $1028\times752$ \\
        \bottomrule
    \end{tabular}
\end{table}

\section{Additional Results}
\subsection{Synthetic Marker Placement}
To stress-test our algorithm, we randomly place markers to simulate different marker layouts. To get marker placement, we uniformly sample the surface based on surface area and add an offset of 9.5mm to the surface of the ground-truth SMPL mesh. We acquire layouts with 20, 30, 40, and 50 markers (see Fig.~\ref{fig:synthetic_num_markers} for reconstructions). We only generate the 50-marker layout and then progressively remove 10 markers to get the other layouts. We do this with 10 different random seeds, effectively producing 10 unique layouts for each number of markers.

\begin{figure}
    \centering
    \begin{subfigure}{0.49\linewidth}
        \includegraphics[width=\linewidth]{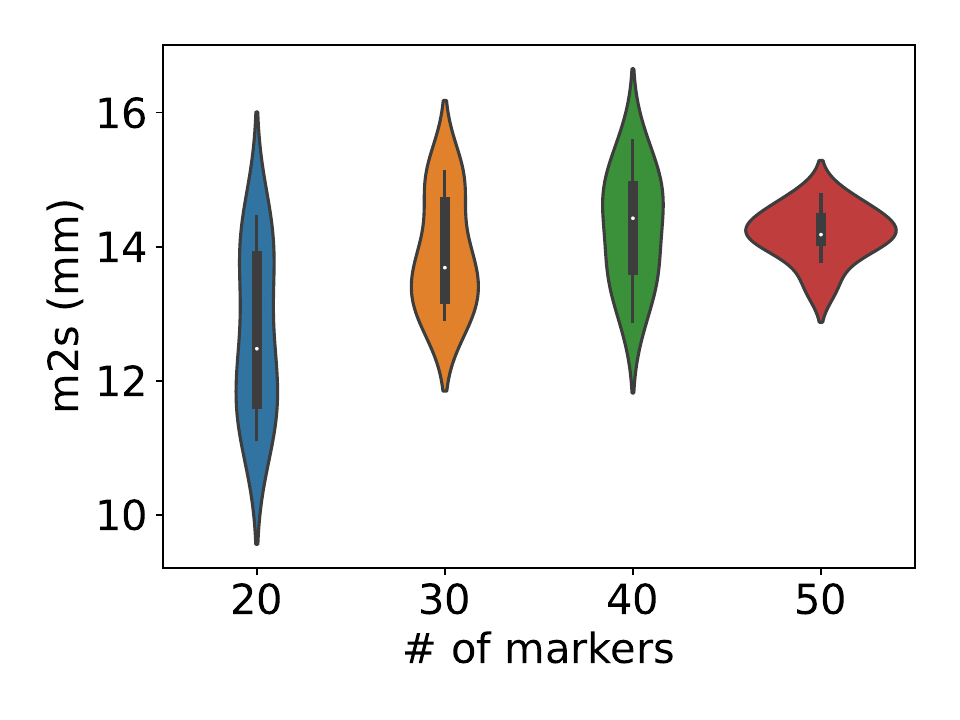}
    \end{subfigure}
    \begin{subfigure}{0.49\linewidth}
        \includegraphics[width=\linewidth]{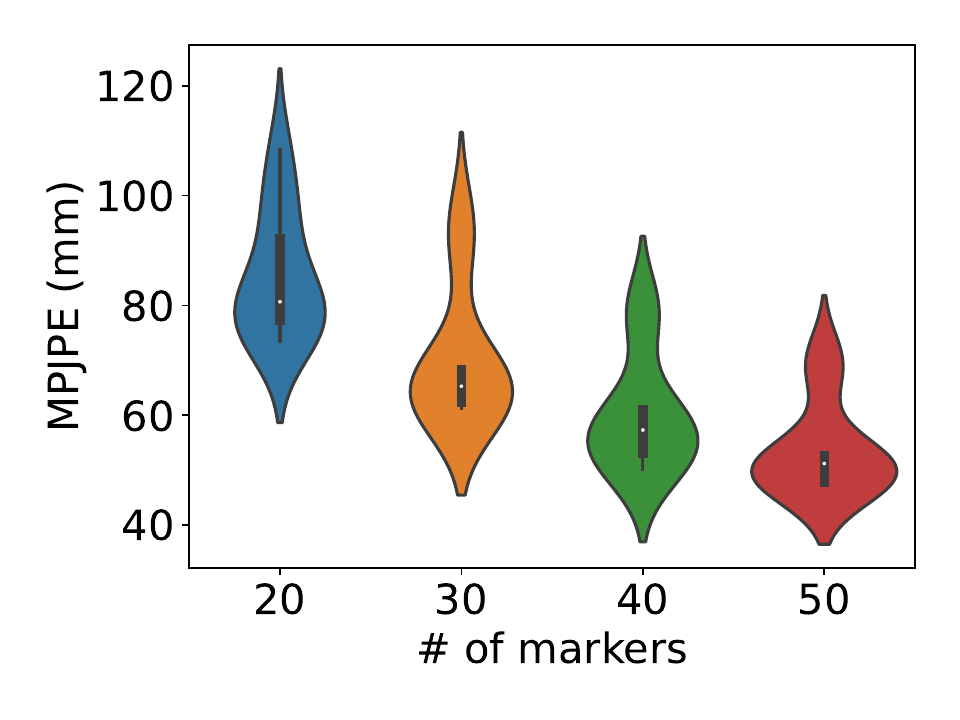}
    \end{subfigure}
    \begin{subfigure}{0.49\linewidth}
        \includegraphics[width=\linewidth]{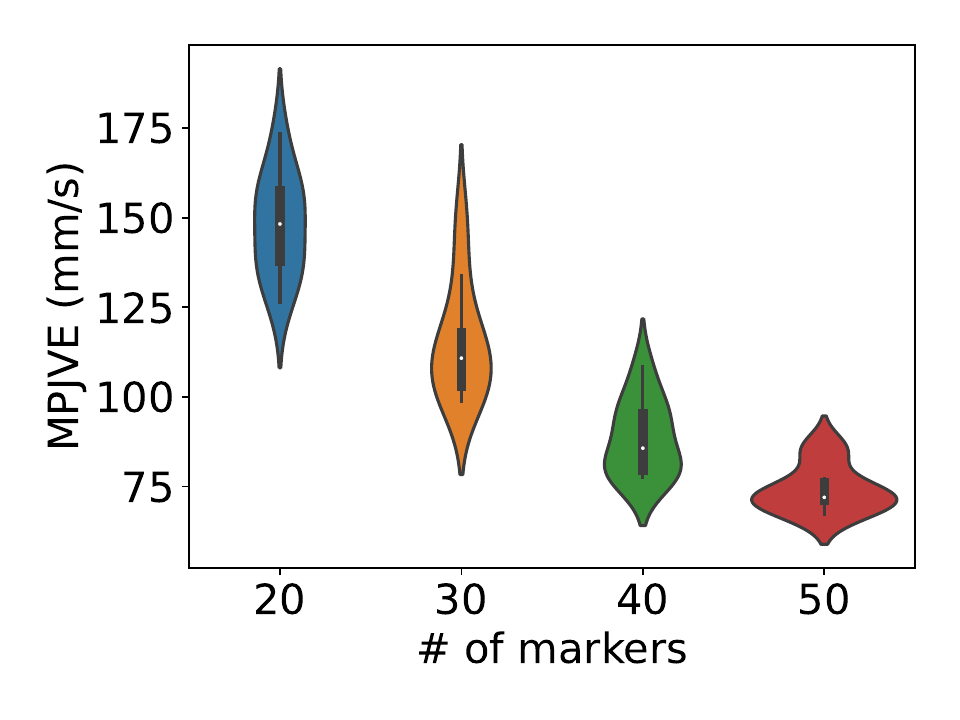}
    \end{subfigure}
    \begin{subfigure}{0.49\linewidth}
        \includegraphics[width=\linewidth]{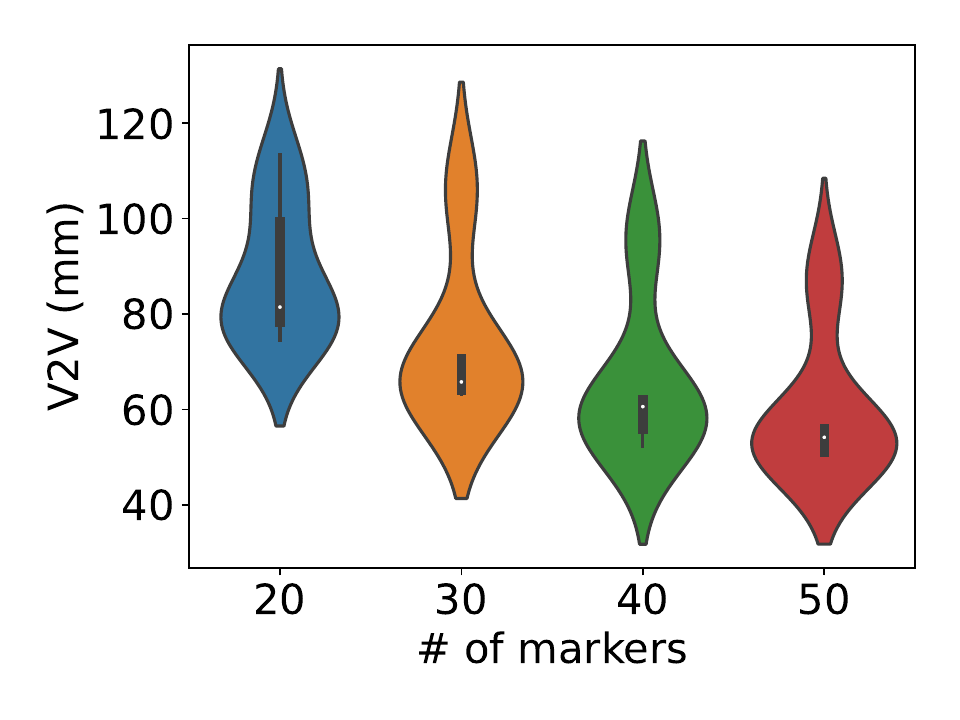}
    \end{subfigure}
    \caption{Synthetic benchmark. We generate and evaluate synthetic examples for 10 different synthetic layouts with 20, 30, 40, and 50 markers. Our method produces lower MPJPE, MPJVE, and V2V with a higher number of markers. As the layouts are randomly generated, the reconstruction error can vary depending on the marker placement.}
    \Description{Synthetic benchmark. We generate and evaluate synthetic examples for 10 different synthetic layouts with 20, 30, 40, and 50 markers.}
    \label{fig:synthetic_plots}
\end{figure}

As seen in Fig.~\ref{fig:synthetic_plots}, we test different numbers of markers to show that our technique generally has lower errors with more markers. Importantly, because the markers are randomly placed, they may not be placed in optimal positions.

\subsection{Video Reconstruction Error Robustness}
While HMR 2.0~\cite{goel2023humans} provides accurate results in general, it does fail in certain cases. For example, we observed problems in reconstruction when self-occlusions are present. Additionally, sometimes tracking can temporarily fail (see Fig.~\ref{fig:tracking_loss}). Our approach generally recovers well from these issues because we mostly rely on the video reconstruction results for initialization.

\subsection{Marker Tracking Loss}
Our method is robust against markers that get lost during tracking (e.g., from occlusions). In our implementation, markers with lost tracking are masked out during optimization, so they only contribute to frames in which they are visible (i.e., the marker position $m^{(t)}\neq (0, 0, 0)$).

To test robustness for marker loss, we simulate marker loss by randomly dropping markers. The results of these experiments are shown in Fig.~\ref{fig:marker_loss}. Each frame, we hide markers with probabilities of (0.0\%, 0.2\%, 1.0\%) and keep them hidden for 10 frames. Even with multiple dropped markers, our approach can still perform accurate mocap solving.

\newcommand{\sswidthhmr}{0.18\linewidth}

\newcommand{\sswidthsynthetic}{0.24\linewidth}

\begin{figure}
    \centering
    \setlength{\tabcolsep}{1pt}
    \begin{tabular}{cccc}
        $|M|=20$ & $|M|=30$ & $|M|=40$ & $|M|=50$ \\
        \includegraphics[width=\sswidthsynthetic]{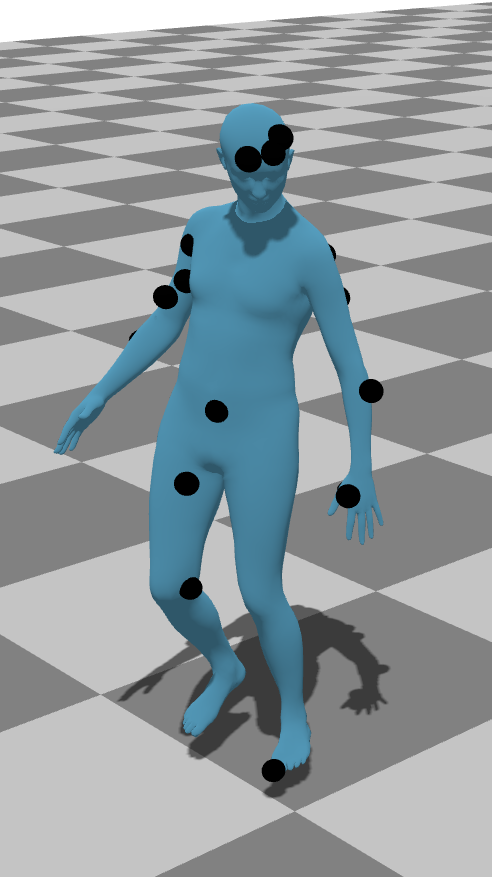} &
        \includegraphics[width=\sswidthsynthetic]{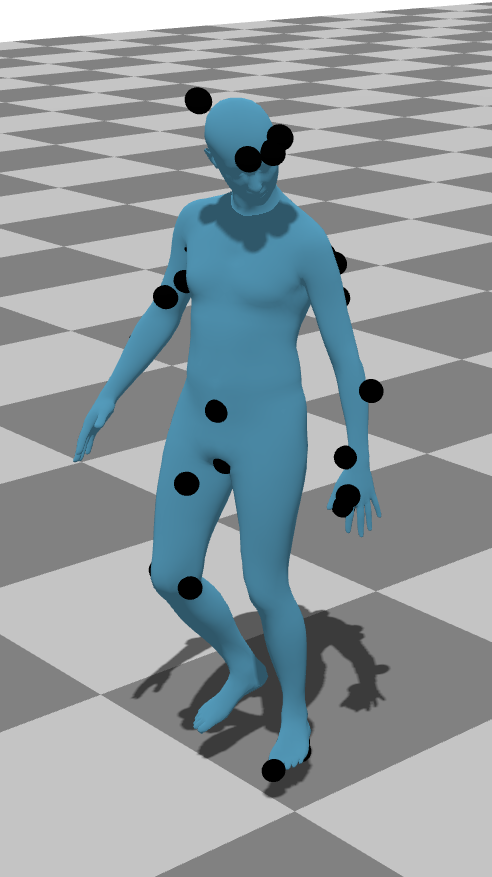} &
        \includegraphics[width=\sswidthsynthetic]{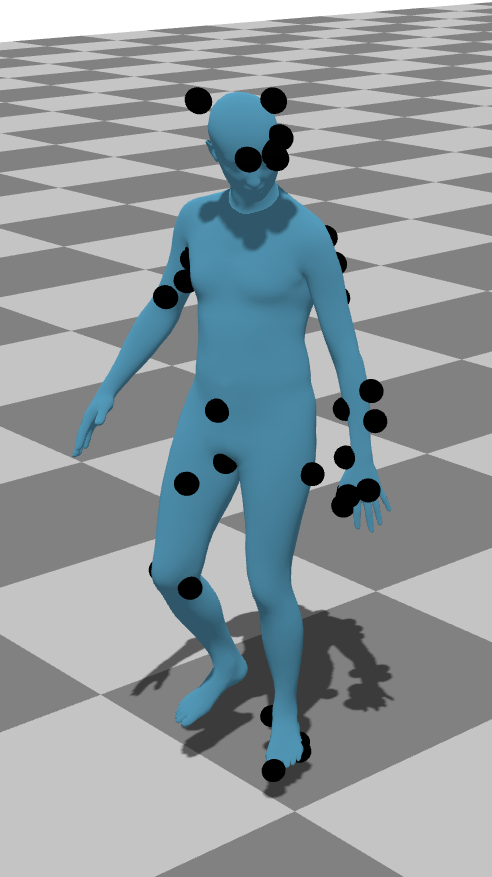} &
        \includegraphics[width=\sswidthsynthetic]{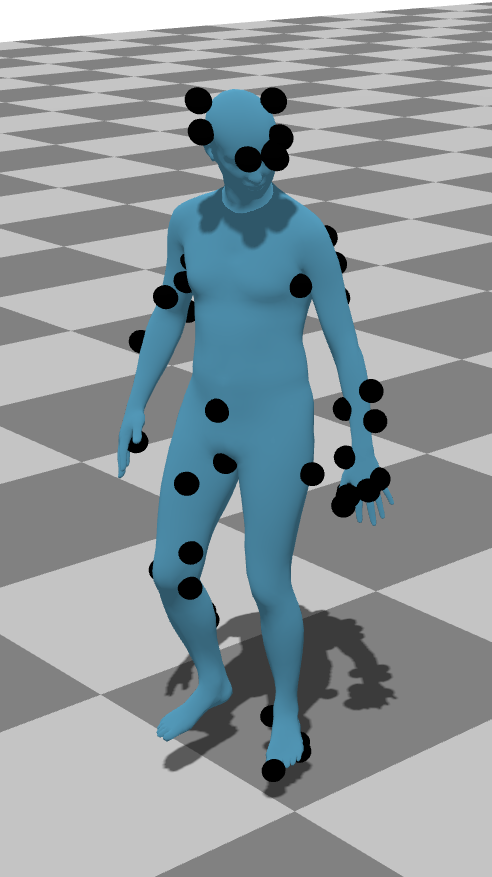}
    \end{tabular}
    \caption{Our reconstruction for synthetic mocap data for layouts with 20, 30, 40, and 50 markers. While our method works better with more markers, it can still solve even with lower numbers of markers.}
    \Description{Our reconstruction for synthetic mocap data for layouts with 20, 30, 40, and 50 markers. While our method works better with more markers, it can still solve even with lower numbers of markers.}
    \label{fig:synthetic_num_markers}
\end{figure}

\begin{figure}
    \centering
    \setlength{\tabcolsep}{1pt}
    \begin{tabular}{cccc}
        \includegraphics[width=\sswidthsynthetic]{images/appendix/synthetic_1_50.png} &
        \includegraphics[width=\sswidthsynthetic]{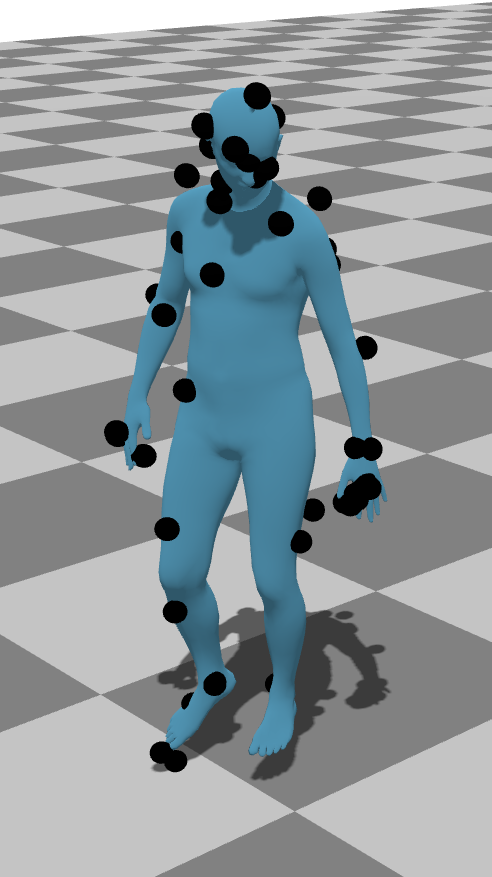} &
        \includegraphics[width=\sswidthsynthetic]{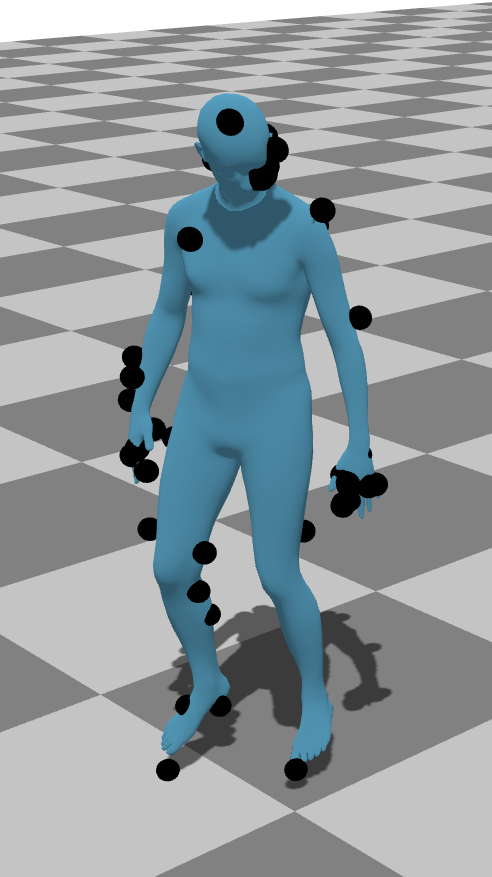} &
        \includegraphics[width=\sswidthsynthetic]{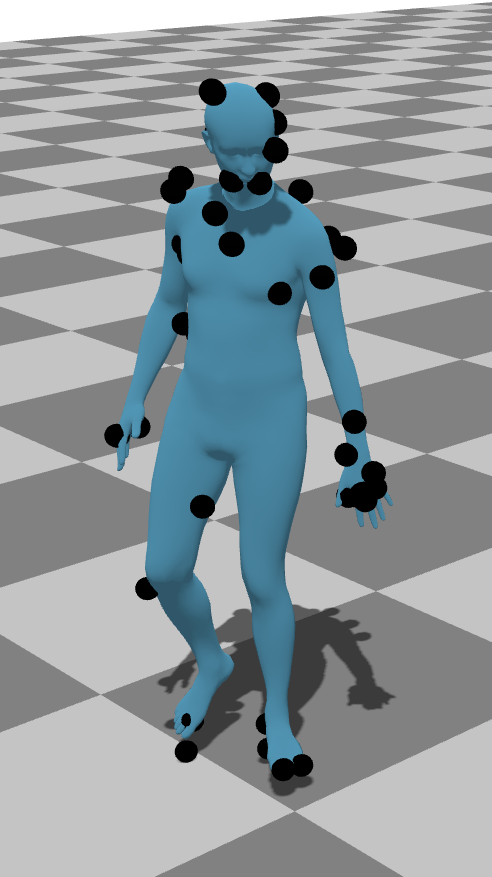}
    \end{tabular}
    \caption{Our reconstruction with different seeds for marker placement with 50 markers. Our method is robust against different marker layouts, even if marker placement is sub optimal.}
    \Description{Our reconstruction with different seeds for marker placement with 50 markers. Our method is robust against different marker layouts, even if marker placement is sub optimal.}
    \label{fig:synthetic_seeds}
\end{figure}

\begin{figure}
    \centering
    \setlength{\tabcolsep}{1pt}
    \begin{tabular}{ccccc}
        60 & 80 & 100 & 120 & 140 \\
        \includegraphics[width=\sswidthhmr]{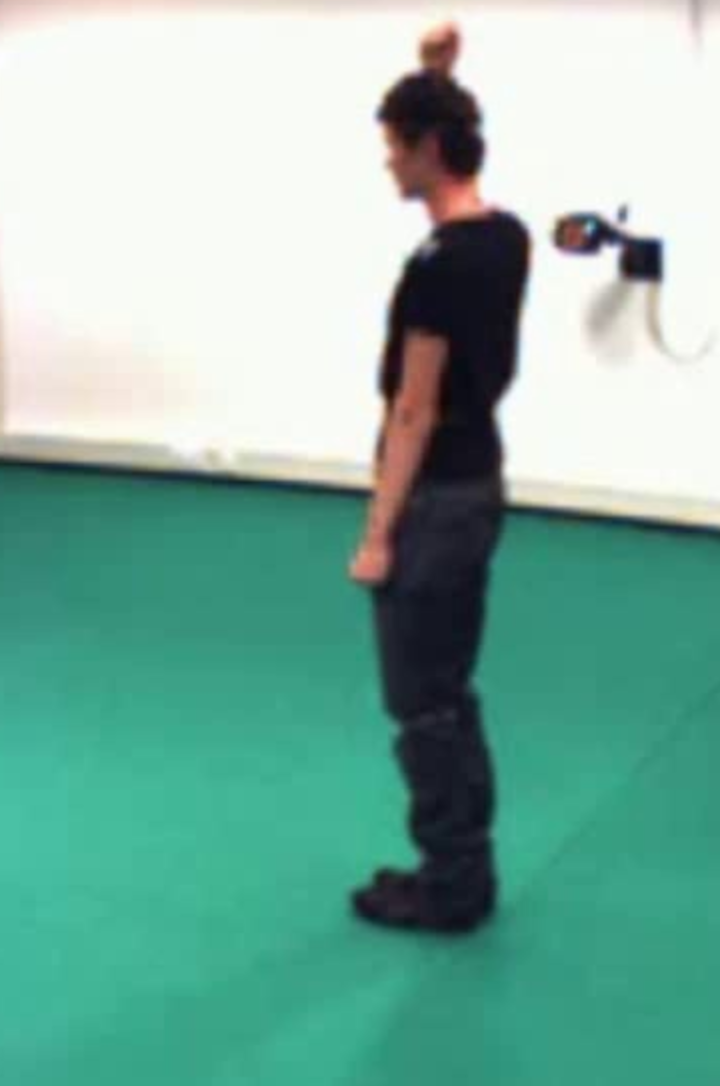} &
        \includegraphics[width=\sswidthhmr]{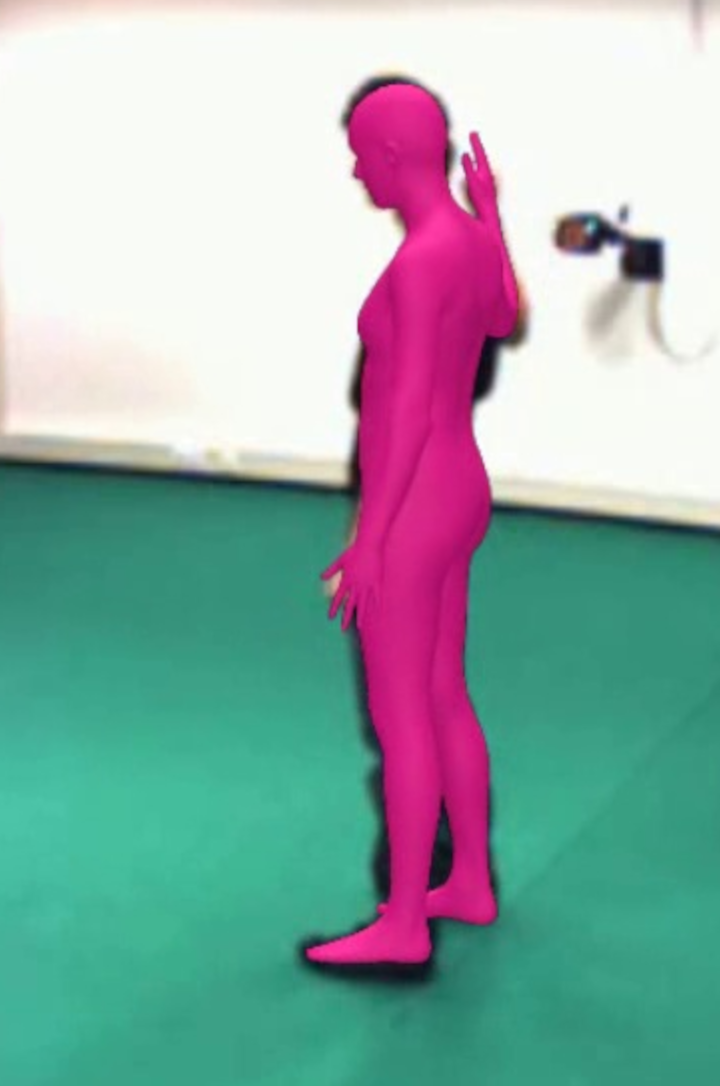} &        
        \includegraphics[width=\sswidthhmr]{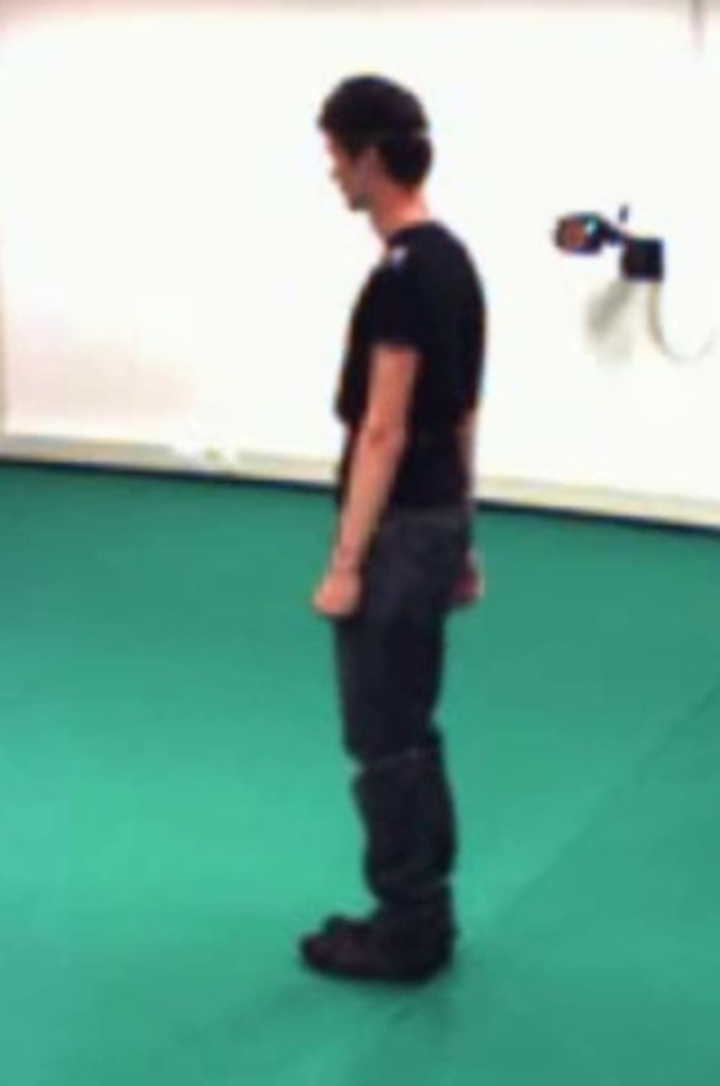} &
        \includegraphics[width=\sswidthhmr]{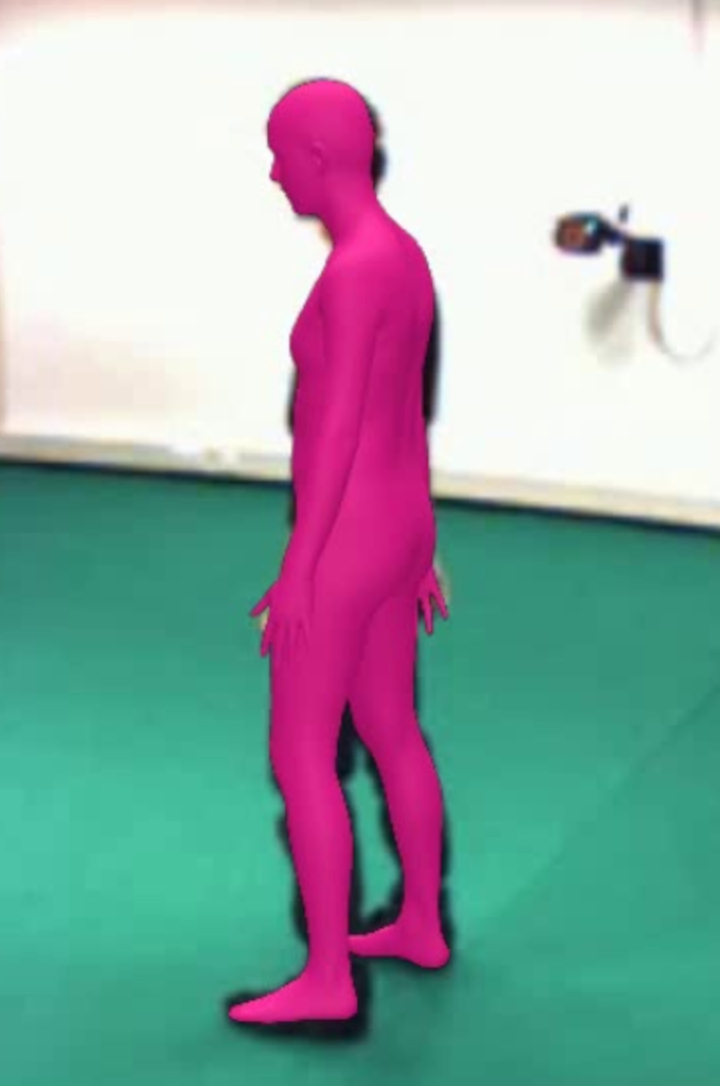} &
        \includegraphics[width=\sswidthhmr]{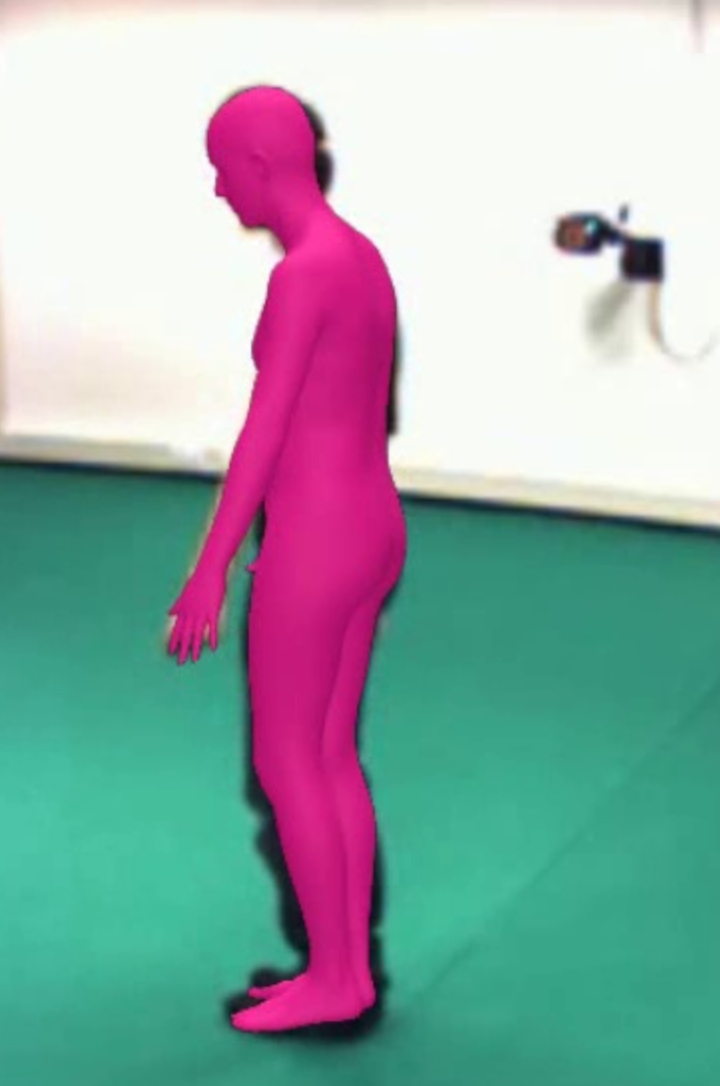} \\

        \includegraphics[width=\sswidthhmr]{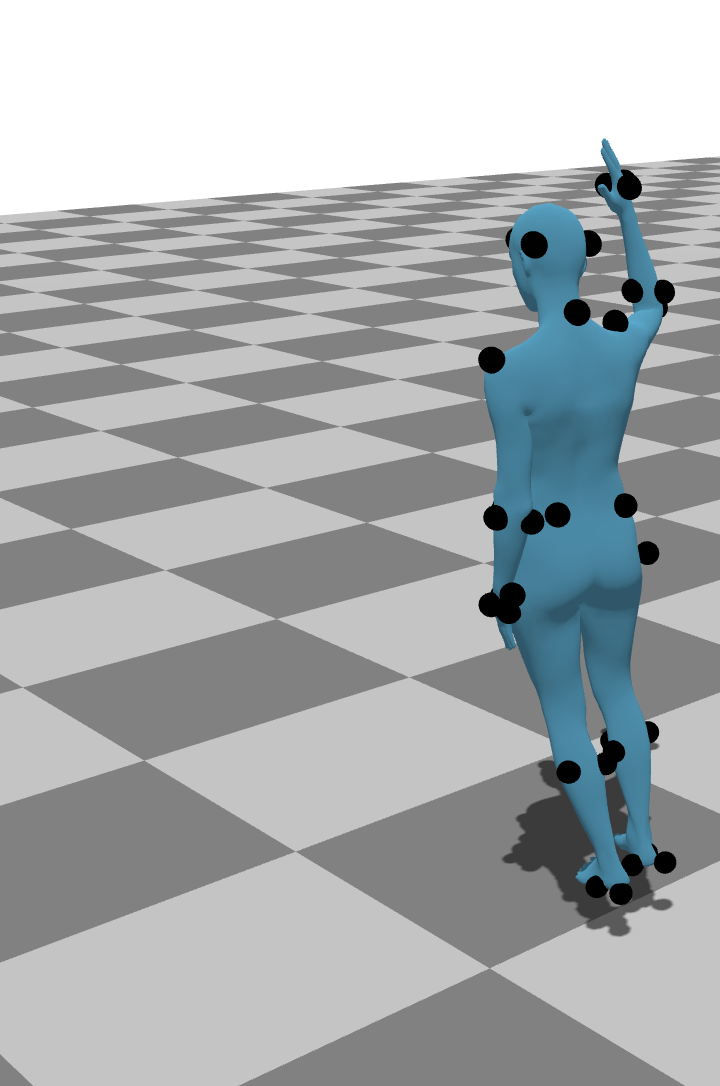} &
        \includegraphics[width=\sswidthhmr]{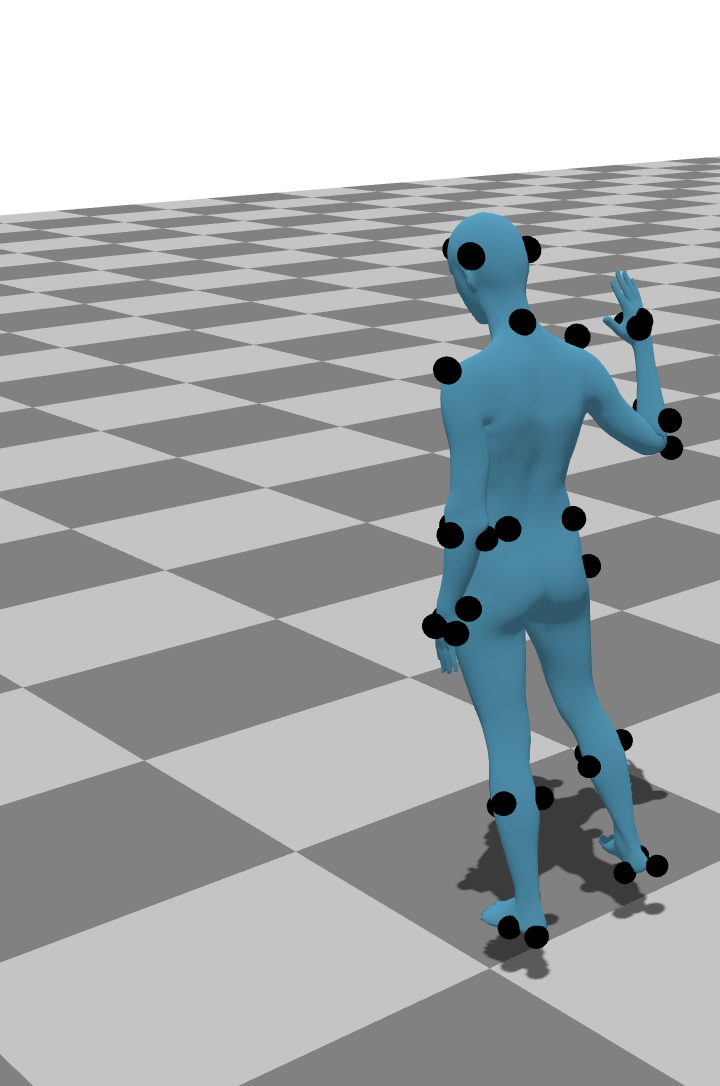} &        
        \includegraphics[width=\sswidthhmr]{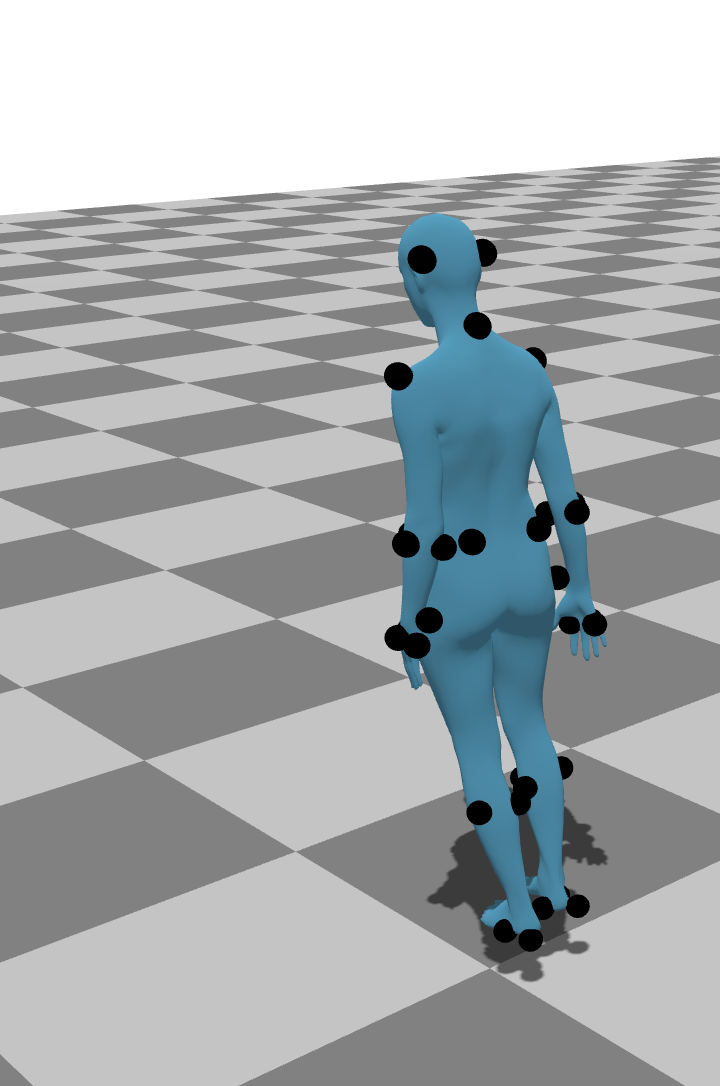} &
        \includegraphics[width=\sswidthhmr]{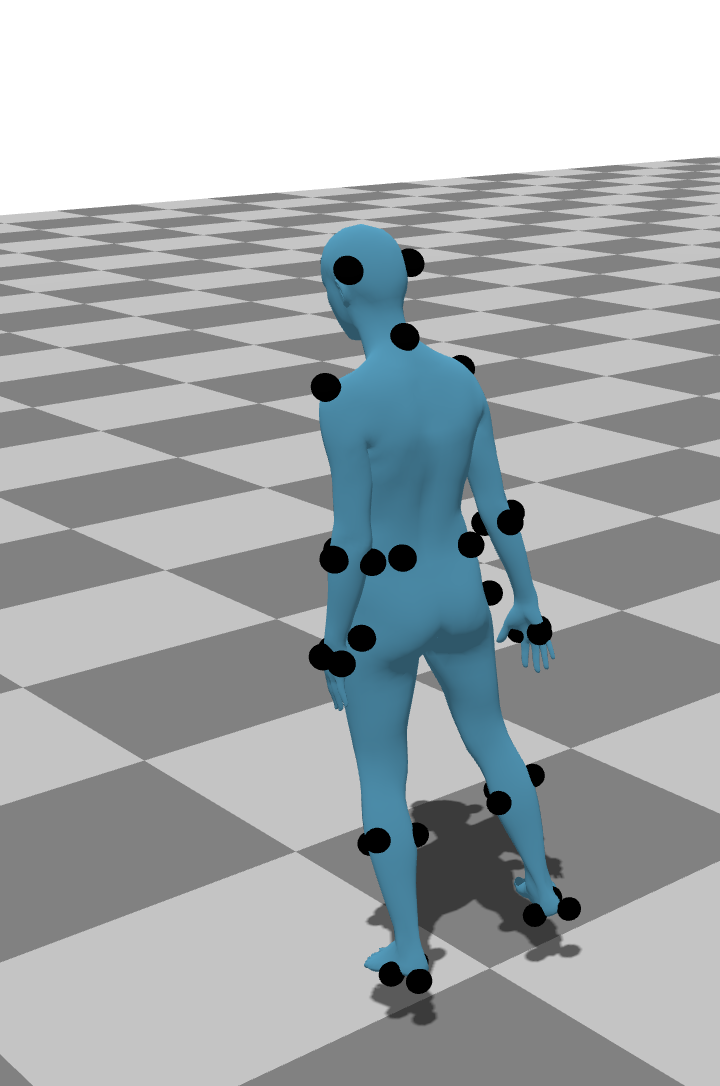} &
        \includegraphics[width=\sswidthhmr]{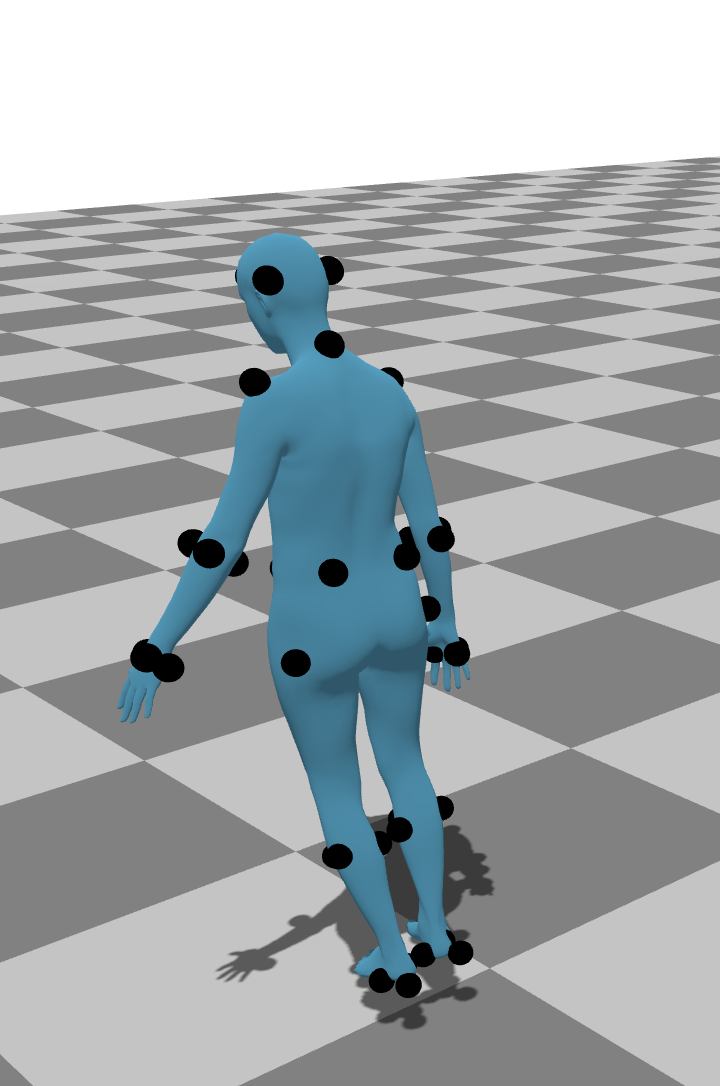} \\
    
        \includegraphics[width=\sswidthhmr]{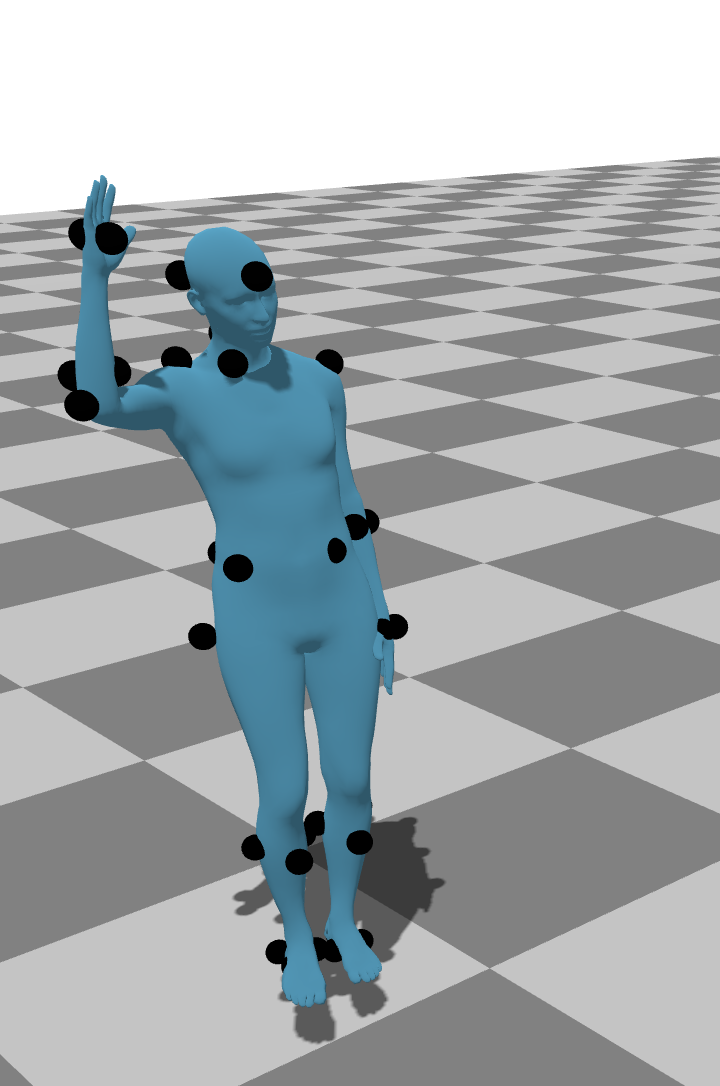} &
        \includegraphics[width=\sswidthhmr]{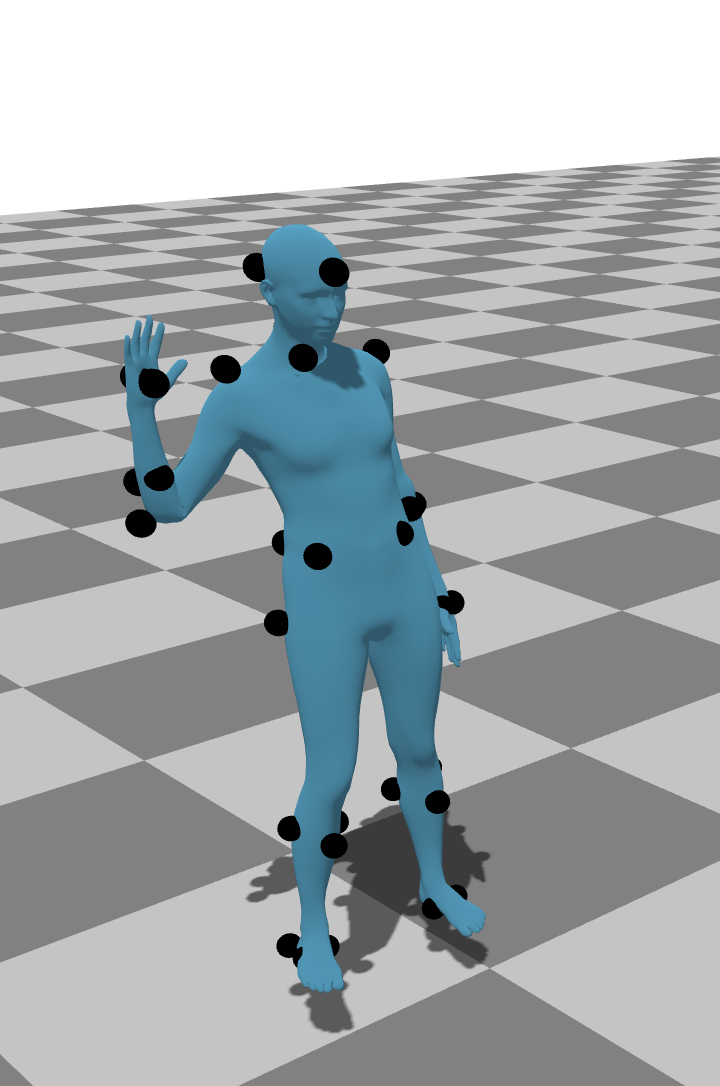} &        
        \includegraphics[width=\sswidthhmr]{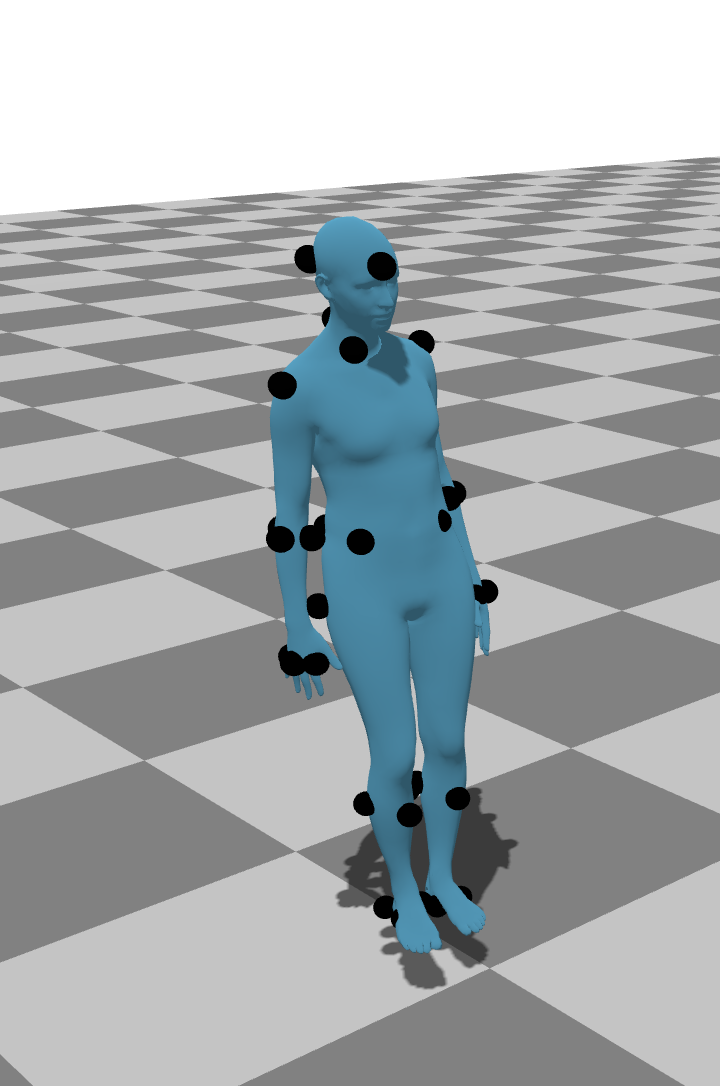} &
        \includegraphics[width=\sswidthhmr]{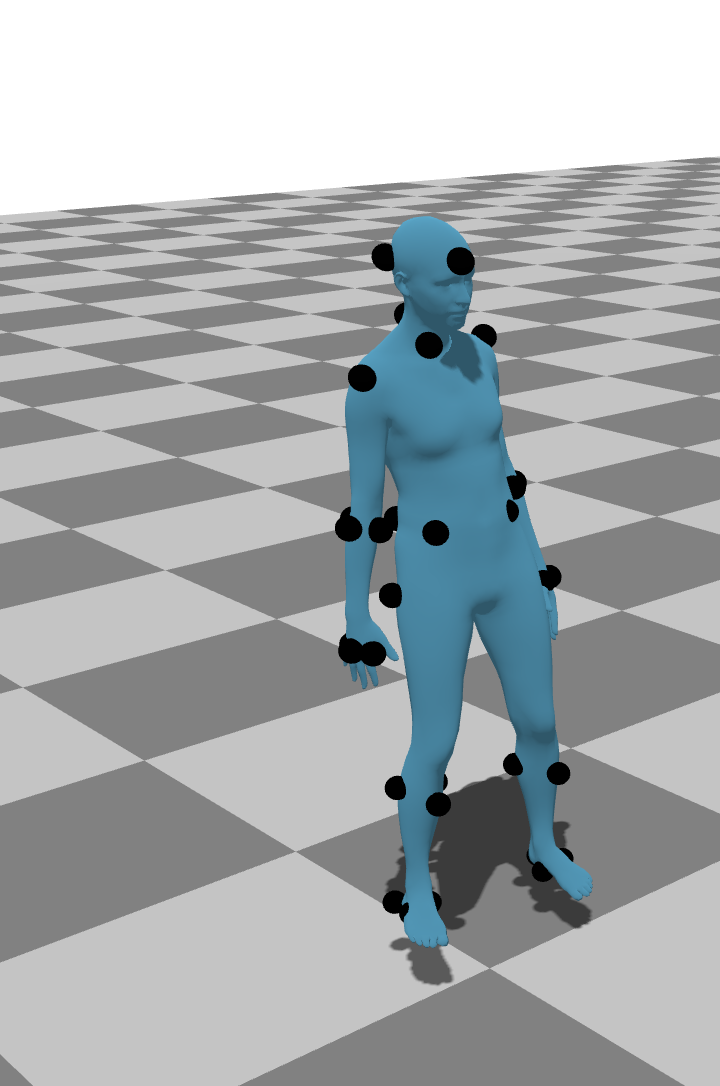} &
        \includegraphics[width=\sswidthhmr]{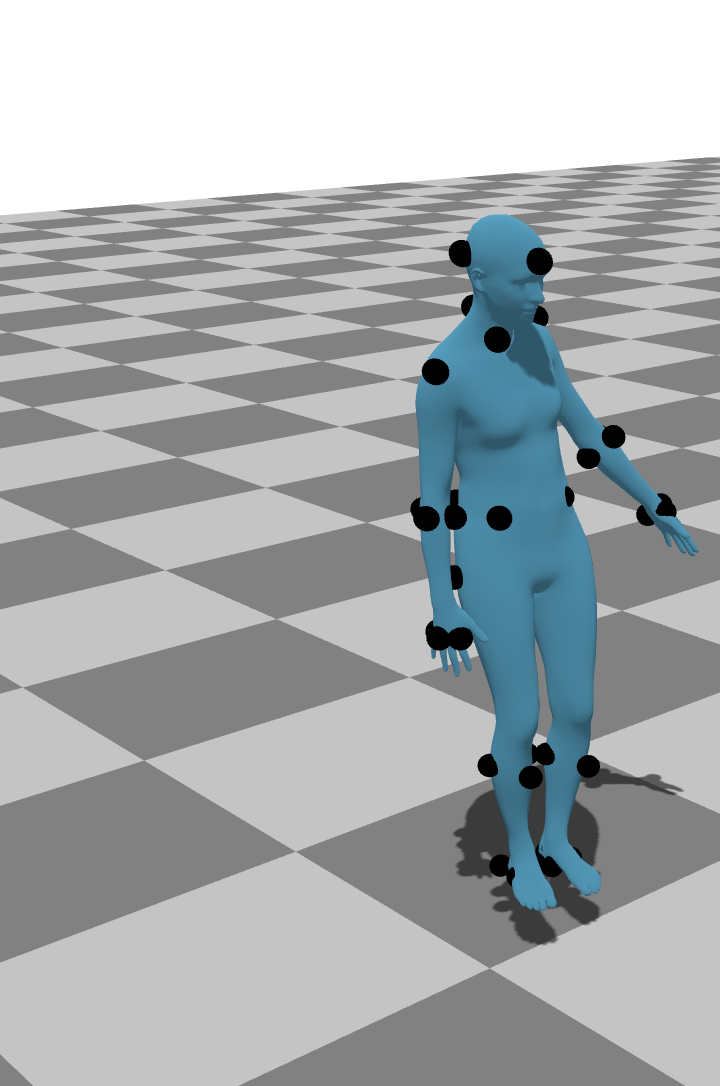} \\
    \end{tabular}
    \caption{\small Qualitative results for a sequence from UMPM~\cite{HICV11:UMPM} at frames \{60, 80, 100, 120, 140\}. First row: the pink reprojected SMPL overlay shows the tracked person by HMR 2.0~\cite{goel2023humans}. Images without the pink overlay show tracking failure. Second and third rows: successful reconstruction with our method from two different camera angles.}
    \Description{Qualitative results for a sequence from UMPM~\cite{HICV11:UMPM} at frames \{60, 80, 100, 120, 140\}. First row: the pink reprojected SMPL overlay shows tracked people by HMR 2.0~\cite{goel2023humans}. Images without the pink overlay show tracking failure. Second and third rows: successful reconstruction with our method from two different camera angles. Our method can recover from failures in monocular human mesh recovery.}
    \label{fig:tracking_loss}
\end{figure}

\newcommand{\mlwidth}{0.32\linewidth}
\begin{figure}
    \centering
    \setlength{\tabcolsep}{1pt}
    \begin{tabular}{ccc}
        0.0\% & 0.2\% & 1.0\% \\
        \includegraphics[width=\mlwidth]{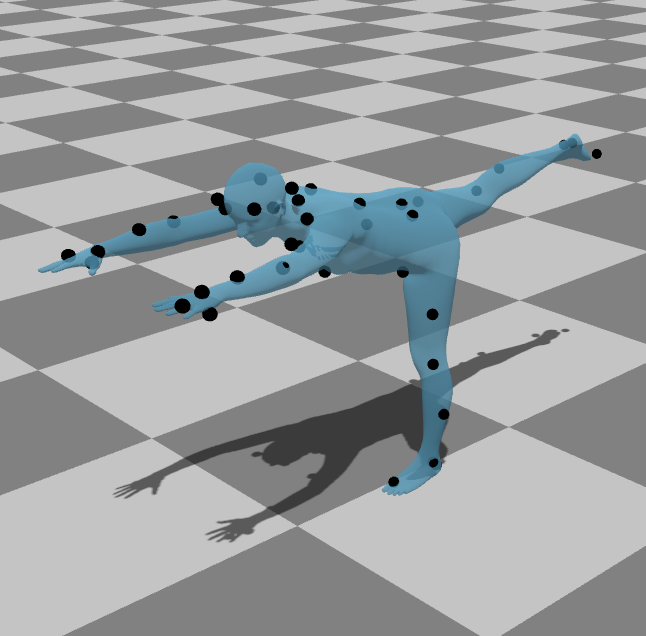} &
        \includegraphics[width=\mlwidth]{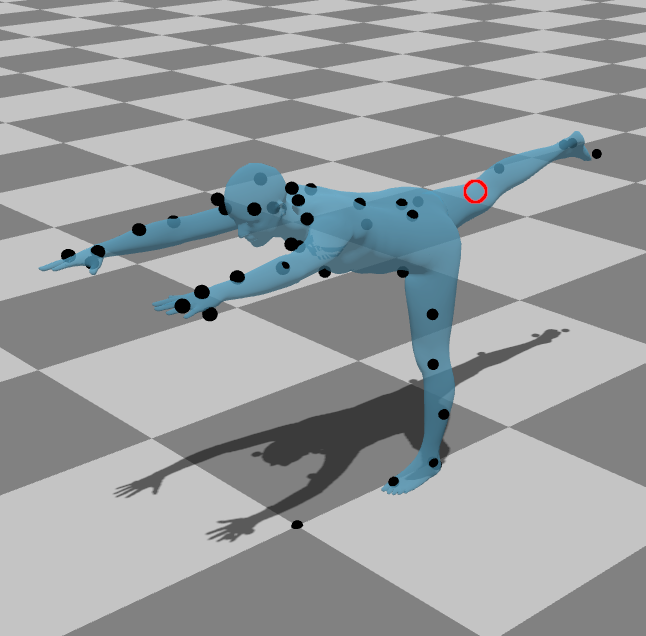} &
        \includegraphics[width=\mlwidth]{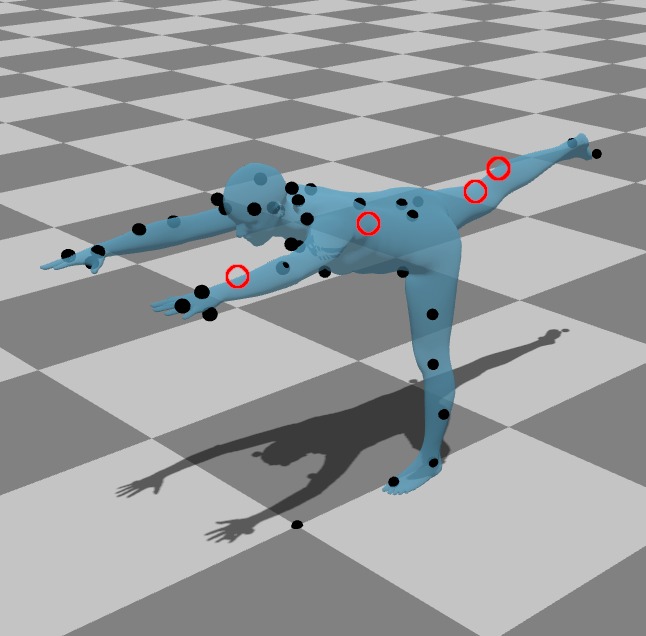} \\
        \includegraphics[width=\mlwidth]{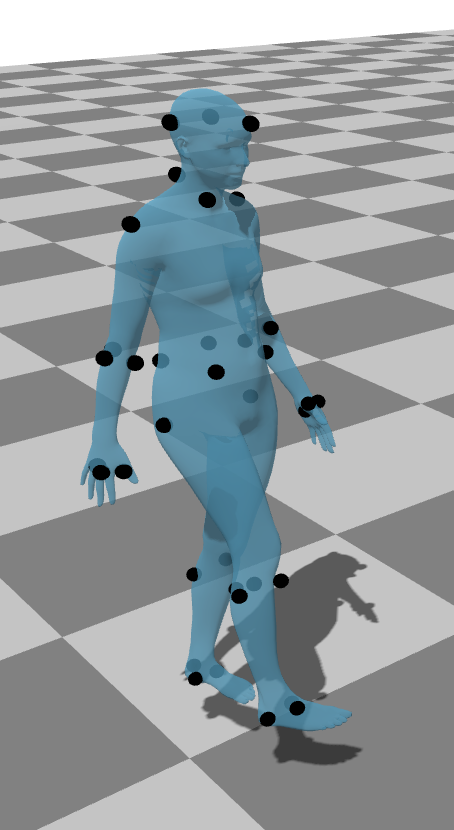} &
        \includegraphics[width=\mlwidth]{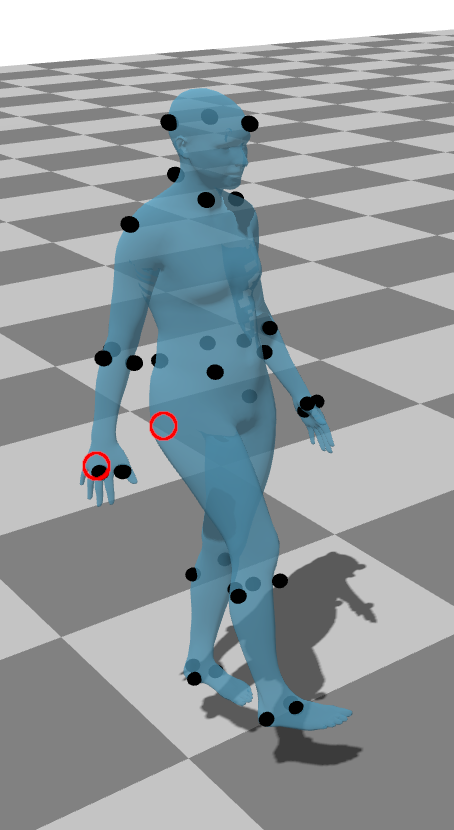} &
        \includegraphics[width=\mlwidth]{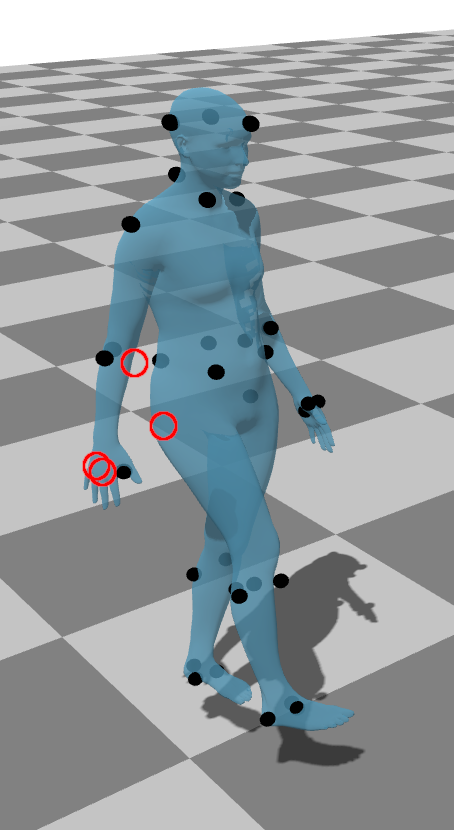} \\
    \end{tabular}
    \caption{\small Reconstruction results for simulated marker loss. The red circles show places where the marker should be located but tracking failed. Even with a few lost markers, our approach can accurately reconstruct results.}
    \Description{Reconstruction results for simulated marker loss. The red circles show places where the marker should be located but tracking failed. Even with a few lost markers, our approach can accurately reconstruct results.}
    \label{fig:marker_loss}
\end{figure}

\let\clearpage\relax
\end{document}